\documentclass{article}

 \usepackage[main, final]{neurips_2026}

\usepackage{microtype}
\usepackage{graphicx}
\usepackage{subcaption}
\usepackage{multirow}
\usepackage{booktabs} 
\usepackage{natbib}
\usepackage{xcolor}
\setcitestyle{numbers}
\setcitestyle{square}
\setcitestyle{comma}

\usepackage{setspace}
\usepackage{longtable}
\usepackage{array}
\usepackage{tabularx}
\usepackage{lipsum} 
\usepackage{wrapfig} 
\usepackage{booktabs} 
\usepackage{bbm}
\usepackage{amsmath}
\usepackage{amssymb}
\usepackage{mathtools}
\usepackage{amsthm}
\usepackage{makecell}
\usepackage{xcolor}         
\usepackage{colortbl}
\usepackage{amsmath}
\usepackage{amssymb}
\usepackage{fontawesome5}
\usepackage{mathtools}
\usepackage{amsthm}
\usepackage{algorithm}
\usepackage{algorithmic}
\usepackage{amsfonts}

\usepackage[utf8]{inputenc} 
\usepackage[T1]{fontenc}    
\usepackage[hidelinks]{hyperref}
\definecolor{citeColor}{RGB}{0,20,115}
\hypersetup{colorlinks,linkcolor={citeColor},citecolor={citeColor},urlcolor={citeColor}}  
\usepackage{url}            
\usepackage{booktabs}       
\usepackage{amsfonts}       
\usepackage{nicefrac}       
\usepackage{microtype}      

\usepackage{amsmath}
\usepackage{array}
\usepackage{amsthm}
\usepackage{amsfonts}
\usepackage{enumerate}
\usepackage{bm}
\usepackage{tabularx}
\usepackage{enumitem}
\usepackage{multirow}
\usepackage{xcolor}
\usepackage{nicefrac}
\usepackage{booktabs}
\usepackage{bbding}
\usepackage{caption}
\usepackage{graphicx}
\usepackage{bbding}
\usepackage{wrapfig}
\usepackage{lipsum}
\usepackage{amssymb}
\usepackage{multibib}
\usepackage{tcolorbox}
\tcbuselibrary{skins,breakable}
\usepackage{booktabs,wasysym}

\newtcolorbox{promptbox}[1]{
  enhanced, breakable,
  colback=gray!5, colframe=black!70,
  colbacktitle=black!75, coltitle=white,
  fonttitle=\small\bfseries,
  title={#1},
  boxrule=0.5pt, arc=1.5pt,
  left=6pt, right=6pt, top=4pt, bottom=4pt,
  fontupper=\small,
}

\newtcolorbox{prompttemplate}[1]{
  enhanced, breakable,
  colback=gray!5, colframe=black!60,
  colbacktitle=black!70, coltitle=white,
  fonttitle=\small\bfseries,
  title={#1},
  boxrule=0.5pt, arc=1.5pt,
  left=6pt, right=6pt, top=4pt, bottom=4pt,
  fontupper=\small\ttfamily,
}

\usepackage[normalem]{ulem}
\usepackage{varwidth}
\usepackage{tikz}
\usetikzlibrary{calc}
\usetikzlibrary{positioning, shapes, fit, backgrounds, decorations.pathreplacing}

\usepackage[capitalize,noabbrev]{cleveref}
\usepackage[textsize=tiny]{todonotes}

\usepackage{etoc}
\etocdepthtag.toc{mtchapter}
\etocsettagdepth{mtchapter}{subsection}
\etocsettagdepth{mtappendix}{none}

\theoremstyle{plain}

\theoremstyle{definition}

\theoremstyle{remark}

\usepackage{pifont}
\newcommand{\cmark}{\ding{51}} 
\newcommand{\xmark}{\ding{55}} 

\newcommand{\AgingBench}{\textsc{AgingBench }}

\usepackage{xcolor}

\definecolor{cNavy}{HTML}{0072B2}
\definecolor{mechCompr}{HTML}{E07070}
\definecolor{mechIntf}{HTML}{5B9BD5}
\definecolor{mechRev}{HTML}{A475C9}
\definecolor{mechMaint}{HTML}{C9883E}

\title{\textcolor{orange}{\faHourglassHalf \, Your Agents Are Aging Too:}  Agent Lifespan Engineering for Deployed Systems}

\author{
	\textbf{Jianing Zhu}$^{1}$\thanks{Equal Contribution} \;
	\textbf{Yeonju Ro}$^{1*}$ \;
	\textbf{John T. Robertson}$^{1}$ \;
    \textbf{Kevin Wang}$^{1}$ \;
    \textbf{Junbo Li}$^{1}$ \;
    \\
    \textbf{Haris Vikalo}$^{1}$\;
    \textbf{Aditya Akella}$^{1}$ \;
	\textbf{Zhangyang Wang}$^{1}$
	\vspace{2mm} \\
	$^{1}$The University of Texas at Austin\\
    \vspace{2mm} \\
    \vspace{-4mm}
}

\begin{document}

\maketitle

\begin{abstract}
Long-lived AI agents are increasingly deployed as persistent operational systems, yet they are still mostly evaluated like freshly initialized models. Day-one benchmarks miss a basic systems question: \textit{how does an agent's reliability change after deployment}? Even when model weights are frozen, an agent's effective state keeps changing as it compresses interaction history, retrieves from a growing memory store, revises facts after updates, and undergoes routine maintenance. Reliability therefore becomes a lifespan property of the full agent harness, not only a snapshot property of the base model.
We introduce \textsc{AgingBench}, a longitudinal reliability benchmark for \textbf{agent lifespan engineering}: measuring not only whether deployed agents degrade, but what form the degradation takes and where repair should target. \textsc{AgingBench} organizes agent \textit{aging} into four mechanisms: \textit{compression aging}, where writing and rewriting drops future-relevant details; \textit{interference aging}, where accumulated similar memories crowd out the target fact; \textit{revision aging}, where changed or derived state is not updated correctly; and \textit{maintenance aging}, where lifecycle events such as model swapping trigger regressions. To diagnose these failures, \textsc{AgingBench} uses temporal dependency graphs and counterfactual probes that produce diagnostic profiles for the write, retrieval, and utilization stages of the memory pipeline. Across 6 scenarios, 10 memory systems and 6 models from two families, we find that \textbf{agent aging is not one-dimensional} over 400 runs spanning 50--200 sessions: a well-formed reply can hide a value the memory no longer supplies; derived state collapses as updates accumulate; and the same wrong answer can require different repairs depending on what the diagnostic profile points to. These results suggest that reliable agent deployment requires lifespan evaluation, mechanism-level diagnosis, and stage-targeted repair, not only stronger day-one models. Our project is publicly available at: \url{https://agingbench.github.io}.
\end{abstract}

\section{Introduction}\label{sec:intro}

The reliability of long-running systems degrades over time:
databases accumulate stale indices~\citep{bates1998indexing}; software accrues technical debt~\citep{ramasubbu2016technical}; human memory fades, distorts, and conflates experiences with age~\citep{disouky2026human}. Memory-enabled AI agents~\citep{chhikara2025mem0,luo2025large,gao2025survey}, systems that persist state across sessions, are subject to the same pressure, which we term \textit{aging}.
Even with frozen weights, their effective system state~\citep{yao2022react} keeps changing as coding agents accumulate workspace context across sessions~\citep{chen2021evaluating,luo2023wizardcoder} and enterprise agents grow project context over months~\citep{zeng2025routine}. These systems also undergo routine operational events: model updates, memory compaction, or prompt revisions.
The decline can be gradual and partly hidden: an agent may keep answering fluently while the exact value that matters has disappeared, a similar entry has taken its place, or an obsolete fact remains active.
Yet unlike databases, which have external analysis tools~\citep{wang2022extensible}, and software, which has regression test suites~\citep{harrold2001regression}, long-lived agents lack an established way to measure and diagnose how their reliability degrades after deployment.


Recent memory benchmarks~\citep{memoryarena,amabench,amemgym,vehiclemembench,beliefshift,perma} have begun to study long-context and multi-session memory and show that agent performance can degrade as context grows.
Most of them, however, still report reliability as an end-to-end score, with little control over how facts depend on, supersede, or compete with facts from earlier sessions, and without localizing where in the agent a failure arises (\S\ref{sec:related}).
Evaluating long-lived agents as \emph{evolving} systems needs more than a score. We refer to this problem space as \textbf{agent lifespan engineering} (ALE): measuring, diagnosing, and repairing degradation in long-running agent systems. ALE asks three questions:

\newtcolorbox{researchquestion}{
  colback=orange!5,     
  colframe=black,      
  boxrule=0.1pt,       
  arc=3pt,             
  left=4pt,
  right=4pt,
  top=4pt,
  bottom=4pt,
}

\begin{researchquestion}
\begin{center}
    \emph{what kind} of degradation is agent aging \textbf{\textit{over time}}? \emph{how} do routine operational events shape it? \\  And \emph{where} in the agent's memory pipeline did it originate?
\end{center}
\end{researchquestion}


\begin{figure}[t]
\centering
\vspace{-2mm}
\includegraphics[width=0.95\columnwidth]{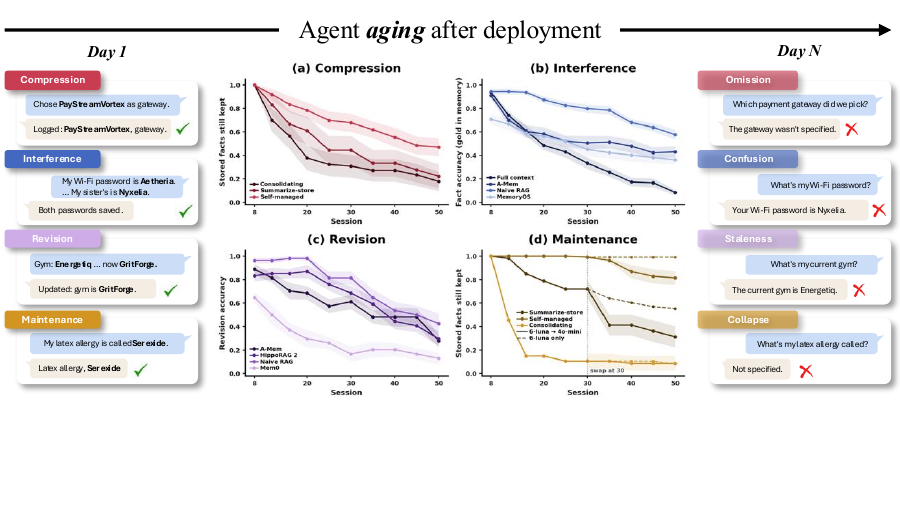}
\vspace{-18mm}
\caption{Four aging mechanisms after deployment. 
(a)~\textit{Compression}: fact retention erodes over sessions as memory rewrites itself. 
(b)~\textit{Interference}: cross-domain recall collapses as accumulated state crowds out the target entries.
(c)~\textit{Revision}: derived-value error compounds across probes when a missed update contaminates downstream answers.
(d)~\textit{Maintenance}: a routine operational event (e.g., replacing the agent internal model) produces divergent recoveries with varied shock types.}
\label{fig:lifecycle}
\vspace{-6mm}
\end{figure}

We introduce \textsc{AgingBench}, a longitudinal reliability benchmark that moves beyond measuring \emph{whether} agents degrade to diagnosing \emph{how}.
We organize agent aging into four mechanisms (refer to Figure~\ref{fig:lifecycle}): \emph{compression aging} (information lost as memory is written and rewritten), \emph{interference aging} (relevant facts buried behind similar entries as state grows), \emph{revision aging} (failure to track changing or derived state), and \emph{maintenance aging} (silent regression after operational events, such as model swapping). To make aging measurable, we introduce a \emph{temporal dependency DAG} (Directed Acyclic Graph) that encodes the cross-session structure: competing content accumulates, values are revised over time, and answers require chaining facts written sessions apart.
Mechanism-specific metrics are then computed based on probes asked at checkpoints.
To simulate longitudinal deployment, all scenarios are backed by \emph{programmatic generators} that scale to multiple sessions, providing a controlled pressure surface for isolating failures that are hard to disentangle in production traces.

A deployed agent is not merely a model but a \emph{harness}: a language model coupled with a memory pipeline~\citep{chhikara2025mem0,xu2026mem,gutierrez2025rag,kang2025memory,liu2026simplemem,fang2026lightmem} for writing, storing, retrieving, and utilizing information.
When the agent fails, ``the memory is getting worse'' is not actionable; knowing \emph{which component} is responsible is.
We therefore build counterfactual conditions into the evaluation: replacing retrieval with an oracle over the agent's own store, and replacing both write and retrieval with the evidence from the history itself.
The resulting profiles point to the stage a repair should target (write, retrieval, or utilization), rather than providing a unique causal decomposition for every architecture.

Across 6 scenarios, 10 memory systems, and 6 models from two families, our experiments (\S\ref{sec:results}) show that agent aging is not one-dimensional.
A well-formed reply can hide a value the memory has lost; derived state collapses as updates accumulate, even when the full history is in context; replacing the model that maintains a memory makes it lose facts it had kept, while retrieval-based and agentic stores are left nearly untouched; and a stronger model repairs only the stages it performs.
The same surface symptom, ``the agent is wrong,'' can therefore point to different failing stages that call for different repairs.
Our contributions can be summarized as follows:
\begin{itemize}[leftmargin=*,nosep]
  \item \textbf{A four-mechanism aging taxonomy} that maps compression, interference, revision, and maintenance to the lifecycle of agent deployment, each with dedicated metrics for auditing (\S\ref{sec:mechanisms}).
  \item \textbf{\textsc{AgingBench}}, a benchmark suite of diverse practical scenarios with programmatic generation and temporal dependency, which support both controlled and autonomous agent evaluation (\S\ref{sec:design}).
  \item \textbf{Component-level attribution} via a configurable evaluation harness that supports counterfactual diagnosis, narrowing ``the agent forgot'' to a specific memory pipeline component: write-time omission, retrieval failure, or utilization failure, identifying the underlying reason of aging (\S\ref{sec:attribution}).
  \item \textbf{Empirical findings} across all four mechanisms showing that agent aging is not one-dimensional: it can hide behind well-formed replies, is sharp for derived state, is triggered by lifecycle events, and depends on the model interaction with its memory architecture (\S\ref{sec:results}).
\end{itemize}

\section{Related Work}\label{sec:related}


Existing work increasingly studies multi-session memory and long-horizon capabilities of AI agents; \textsc{AgingBench} differs by treating \emph{longitudinal degradation itself} as the evaluation object, instrumented through aging curves, a temporal dependency DAG, lifecycle event injection, and component-aware diagnostic profiles. We expand this comparison with detailed discussion in Appendix~\ref{app:related}.



\textbf{Aging in deployed agents.} In practice, long-lived agents face pressures that no snapshot benchmark captures.
A coding agent that compresses months of project context into a fixed-size summary inevitably loses low-frequency details like specific API versions or configuration values~\citep{ge2025survey}.
An enterprise assistant managing multiple clients can retrieve the wrong client's budget when similar entries accumulate in its memory store~\citep{zhu2025compliance}.
A personal planner that once tracked a user's dietary restriction fails to update when the user lifts it, continuing to enforce an obsolete constraint~\citep{yang2026plugmem}.
And a production agent that behaves reliably for weeks silently regresses after a memory recompaction~\citep{yang2025learning}.
Complementary to other benchmarks that evolve the external target (e.g., codebase evolution~\citep{deng2026evoclaw}), our work measures degradation of the agent's internal memory state, with component attribution.
On the memory systems side,
some works~\citep{tiermem,dmem} characterize compression as a bottleneck but do not measure how it degrades agent reliability, nor do they track the full range of deployment pressures.

\textbf{Lifecycle events and attribution for system harnessing.}
Few existing benchmarks (we summarized in Table~\ref{tab:benchmark-landscape}) treat operational events as controlled experimental conditions, and generally assumes a static evaluation environment; the agent memory does not evolve during the benchmark run.
Yet deployed agents routinely undergo such events like memory compacting or flush~\citep{hu2025memory}, and their impact on reliability is rarely measured.
Similarly, failure attribution remains largely unaddressed: existing benchmarks report end-to-end scores without localizing whether the failure originated at write time, during retrieval, or at utilization.
TierMem~\citep{tiermem} partially addresses this by distinguishing summary-caused omissions from reasoning failures, but does not provide a general counterfactual framework.
Our approach adapts counterfactual analysis to inspect the failure of long-lived agents.

\section{Aging Taxonomy in Agent Lifelong Deployment}\label{sec:mechanisms}

We organize agent aging into four mechanisms (Figure~\ref{fig:lifecycle}).
Conceptually, they fall into two families.
\emph{Accumulation-driven aging} (compression, interference) worsens as the agent's state grows over sessions; it is the cost of operating over time, though discrete spikes (e.g., a file overwrite) can punctuate the trend.
\emph{Event-driven aging} (revision, maintenance) is triggered by discrete changes in the environment or agent itself; it is the cost of operating in a world that does not stand still.

\begin{itemize}[leftmargin=*,nosep,itemsep=4pt]
  \item \textbf{Compression aging} arises from the \emph{write-before-query barrier}: memory systems must decide what to preserve at write time, but which facts matter depends on future queries that have not yet arrived~\citep{tiermem,dmem,amabench}. Summarization, consolidation, fact extraction, and self-managed notes can drop or blur specific values and every further rewrite as the history grows can lose more.
  \item \textbf{Interference aging} arises even when no information is lost \emph{and no facts have changed}: as state grows, similar content accumulates (other owners' values, same-topic rounds, look-alike chains, near-miss items) and crowds the evidence out of what the reader sees, or leads it to answer with a competitor~\citep{liu2024lost}. Interference is orthogonal to revision (freezing all facts does not prevent it).
  \item \textbf{Revision aging} occurs when a value changes and the agent answers with a superseded version. A harder form is \emph{derived state}, such as what a value was before or how many times it changed, which must be recomputed from the whole update history.
  \item \textbf{Maintenance aging} occurs when lifecycle events, such as replacing the model that maintains the memory or flushing and migrating stored memory~\citep{wu2016rethinking}, remove knowledge the agent already had. Unlike the other three mechanisms, it is driven by actions taken \emph{on} the agent.
\end{itemize}

\textbf{Deployment scenarios.}
Different agent deployments naturally encounter different subsets of these mechanisms.
A research literature agent accumulates paper summaries over months; published findings rarely change, so its risk lies in losing the details of older papers and confusing similar results.
A lifestyle assistant tracks a user's budgets, diet, and schedule, which change over time, so an outdated preference can look as valid as the current one.
An enterprise knowledge base spans many similar projects whose decisions keep shifting, so the answer for one project can be crowded out by another project's or by an earlier decision.
A production agent also goes through operational events such as routine model rotations, which can remove what it already knew regardless of its domain. Full archetype mapping are discussed in Appendix~\ref{app:curation}.
All four mechanisms can co-occur over an agent's operational lifetime, and the four-way split matters as the same surface symptom ``the agent is wrong,'' needs different interpretations depending on which mechanism is binding.

\begin{figure}[t]
\centering
\vspace{-3mm}
\includegraphics[width=0.94\columnwidth]{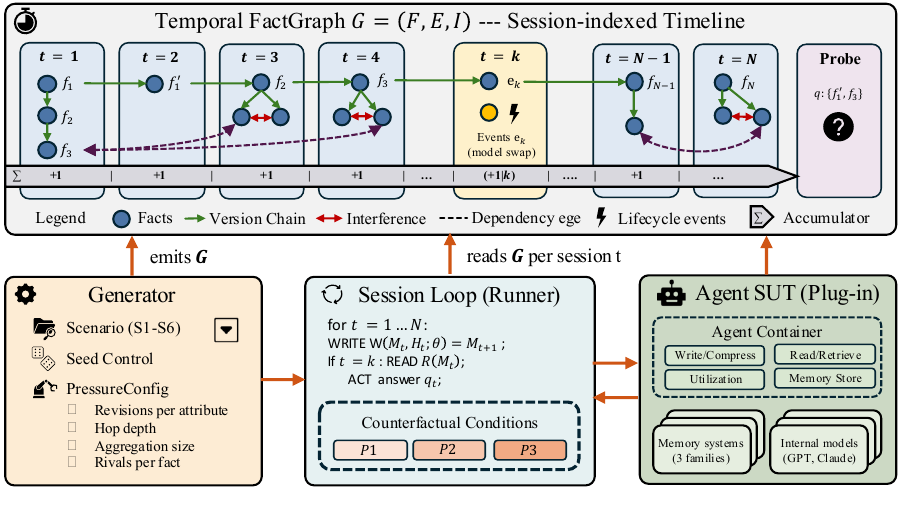}
\vspace{-4mm}
\caption{AgingBench evaluation pipeline.
\textit{Top}: temporal FactGraph as a session-indexed timeline, with version chains, interference rivals, dependency edges, aggregation $\Sigma$, and probes.
\textit{Bottom}: generator emits $G$ and the probe stream; session loop writes every session and probe at checkpoints, with counterfactual conditions; the SUT plugs in a memory system and its internal model.}
\label{fig:pipeline}
\vspace{-5mm}
\end{figure}

\section{\AgingBench: Longitudinal Reliability Benchmark}\label{sec:design}
Making the four aging mechanisms from \S\ref{sec:mechanisms} measurable requires an evaluation framework that can simulate multi-session deployment, encode cross-session dependencies, and scale to long operational lifetimes.
We describe the generation framework that produces cross-session timeline structure at varied scale (\S\ref{sec:dag}) and the evaluation procedure with findings preview (\S\ref{sec:protocol}), in our \textsc{AgingBench}.

\subsection{Timeline Generation with Temporal Structure}\label{sec:dag}

In real deployment, facts accumulate across sessions, supersede each other, and compete for retrieval.
Capturing this structure is essential for making aging measurable: without it, the evaluation cannot distinguish a lost fact from a crowded-out or a stale one.

\textbf{Temporal dependency DAG.}
To encode this cross-session structure, the generator produces a DAG $G = (\mathcal{F}, \mathcal{E}, \mathcal{I})$ for each timeline, in which every planted round carries the session that writes it and the answer key of each checkpoint is computed by graph queries.
\emph{Facts} $\mathcal{F}$ are (subject, relation, value) triples, and \emph{version chains} track supersession within $\mathcal{F}$: when a value is revised, $f_i^{(v)} \to f_i^{(v+1)}$ extends a chain, and the gold at a checkpoint is the latest version written before it.
\emph{Dependency edges} $\mathcal{E}$ link a probe to the facts its answer requires, either a multi-hop chain of depth $d$ or the members of an aggregation.
\emph{Interference rounds} $\mathcal{I}$ attach competitors to a probe: other owners' values for the same attribute, same-topic rounds without a value, look-alike chains, and near-miss items.
These structures generate four probe families as listed in Table~\ref{tab:scenarios}.
\begin{wrapfigure}{r}{0.48\textwidth}
\vspace{-10pt}
\centering
\includegraphics[width=\linewidth]{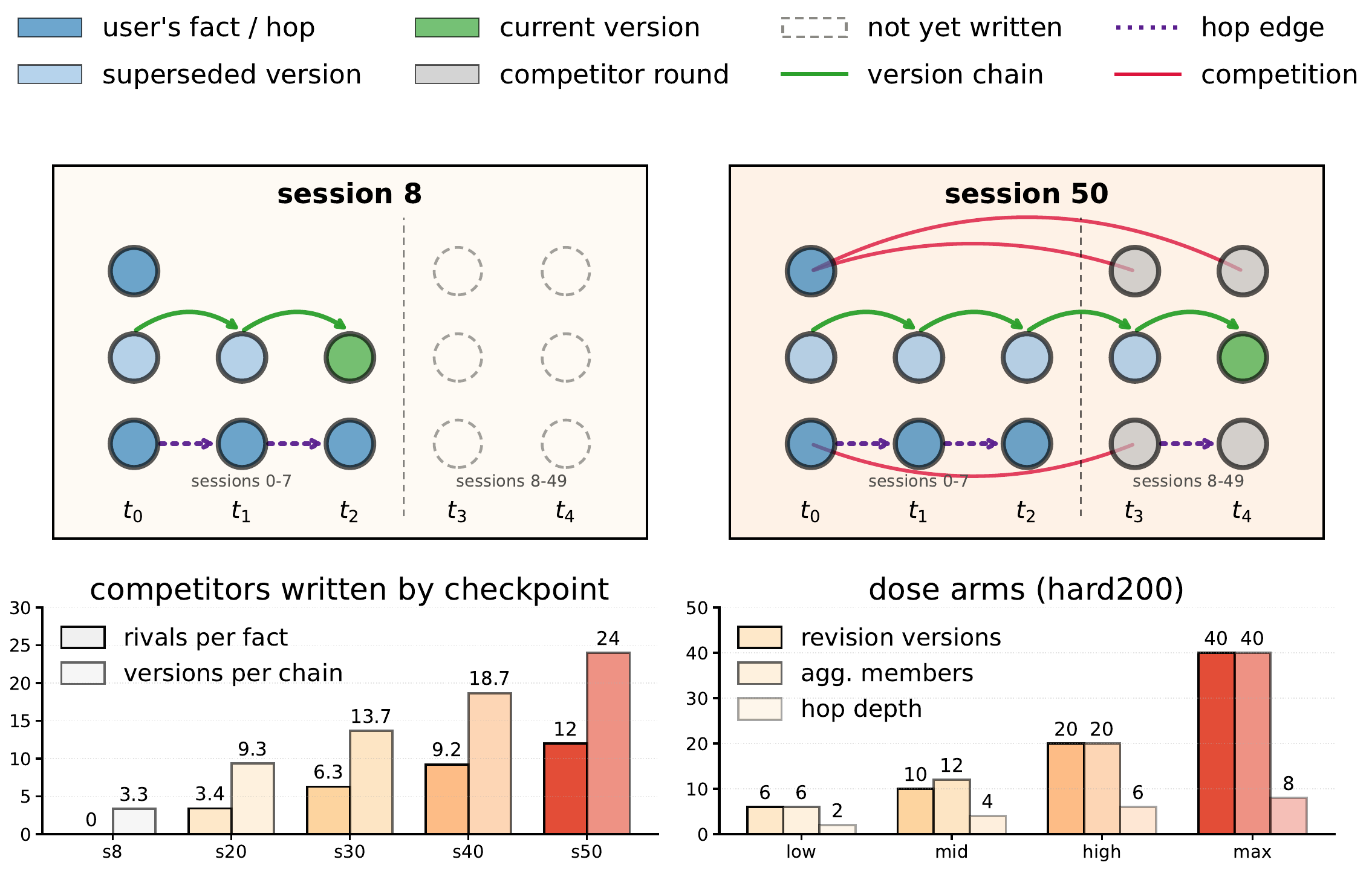}
\vspace{-6mm}
\caption{Temporal DAG and pressure control. Top: one timeline at sessions 8 and 50, with the user's facts, version chains, and competitor rounds. Bottom left: competitors written before each checkpoint in the 50-session builds. Bottom right: the dose arms of the 200-session builds.}
\label{fig:dag}
\vspace{-14pt}
\end{wrapfigure}
\emph{Lookup} asks for one fact in $\mathcal{F}$, with its competitors written later.
\emph{Compose} follows a multi-hop chain in $\mathcal{E}$ to its second or later hop, with a look-alike chain from another root.
\emph{Update} asks for the current value of a version chain, alongside never-updated controls of the same question form.
\emph{Count} aggregates over members in $\mathcal{E}$, with near-miss items and same-topic rounds.
Figure~\ref{fig:dag} illustrates these structures and the pressure they apply.

\textbf{Scalable programmatic generation.}
Measuring aging over long operational lifetimes requires timelines that scale without manual authoring.
\textsc{AgingBench} separates authoring from generation. An offline, LLM-assisted step builds a frozen library of content units for each scenario in Table~\ref{tab:scenarios} and repairs it against an item-level audit (Appendix~\ref{app:validity}).
Generation is deterministic then: it binds units to timelines without sharing a unit across timelines, schedules every edge, and inserts each one as a round into scenario-themed base conversations, so a configuration reproduces its build byte for byte.
Timelines scale to about 240 sessions with the current pools, and pressure knobs (timeline length, competitors per fact, versions per attribute, hop depth, aggregation members) can be varied one at a time, binding the same units across arms where possible.
More details can be found in the appendix: generator and pressure configuration (Appendix~\ref{app:generators}), memory systems and prompts (Appendix~\ref{app:memory}).

\begin{table}[ht]
\centering
\small
\vspace{-4mm}
\caption{Scenario design and probe profile. All scenarios share one planted dose and differ in domain content. Symbols rank each probe family's accuracy drop across scenarios (\CIRCLE{} high, \LEFTcircle{} medium, \Circle{} low). Concrete task examples are provided in Appendix~\ref{app:task-examples}; the scenario summary is in Appendix~\ref{app:scenario-summary}.}
\vspace{2mm}
\label{tab:scenarios}
\resizebox{\textwidth}{!}{
\begin{tabular}{llccccc}
\toprule
\textbf{Scenario} & \textbf{Domain} & \textbf{Lookup} & \textbf{Compose} & \textbf{Update} & \textbf{Count} & \textbf{Words / session} \\
\midrule
S1 Research       & Papers + experiments     & \LEFTcircle & \LEFTcircle & \Circle     & \CIRCLE     & 1972 \\
S2 Lifestyle      & Budget, diet + schedule  & \LEFTcircle & \Circle     & \CIRCLE     & \LEFTcircle & 1702 \\
S3 Knowledge base & Project decisions        & \CIRCLE     & \CIRCLE     & \CIRCLE     & \Circle     & 1805 \\
S4 Software       & Code planning            & \CIRCLE     & \CIRCLE     & \LEFTcircle & \CIRCLE     & 2030 \\
S5 Notebook       & Personal + project notes & \Circle     & \LEFTcircle & \Circle     & \LEFTcircle & 1801 \\
S6 Multi-domain   & Travel, shopping + email & \Circle     & \Circle     & \LEFTcircle & \Circle     & 1695 \\
\bottomrule
\end{tabular}}
\end{table}
\vspace{-2mm}

\subsection{Evaluation Procedure and Aging Preview}\label{sec:protocol}

We formalize agent aging evaluation as a \emph{session loop} over the $N$ sessions of a generated timeline (Figure~\ref{fig:pipeline}), with each memory system plugged in as its own $U$.
At each checkpoint session $t$, the agent reads its memory $M_t$ without changing it, answers the held-out probes $q_t$, and receives a rule-based accuracy score $s_t$.
Every session's dialogue $H_t$ is written into the next state:
\begin{equation}\label{eq:memory-update}
  M_{t+1} = U(M_t, H_t; \theta)
\end{equation}
where $U$ is the memory system's write function ($\mathcal{W}$ in \S\ref{sec:loci}) and $\theta$ its parameters (internal model, prompts, and, where used, a word budget).
At a designated maintenance session $t = k$, the runner injects a lifecycle event $e_k$ that changes $\theta$ by replacing the internal model; a paired run without the event serves as the control, enabling controlled measurement.
The accuracy of a fixed probe set at each checkpoint forms the \textit{aging curve} of a probe family.
A key design principle is \emph{gold-grounded scoring}: every probe is scored by rule against its answer key, with no LLM judge, and each loss is tied to one mechanism.
Compression is the share of facts no longer in memory at the last checkpoint; interference is the accuracy drop on facts whose gold is still in memory; revision is the share of revised attributes answered with a superseded version; maintenance is the additional loss of stored facts caused by a lifecycle event, relative to the paired control.
The full definitions are in Appendix~\ref{app:scoring-defs}.

\begin{figure*}[t!]
    \centering
    \vspace{-2mm}
    \includegraphics[width=0.98\linewidth]{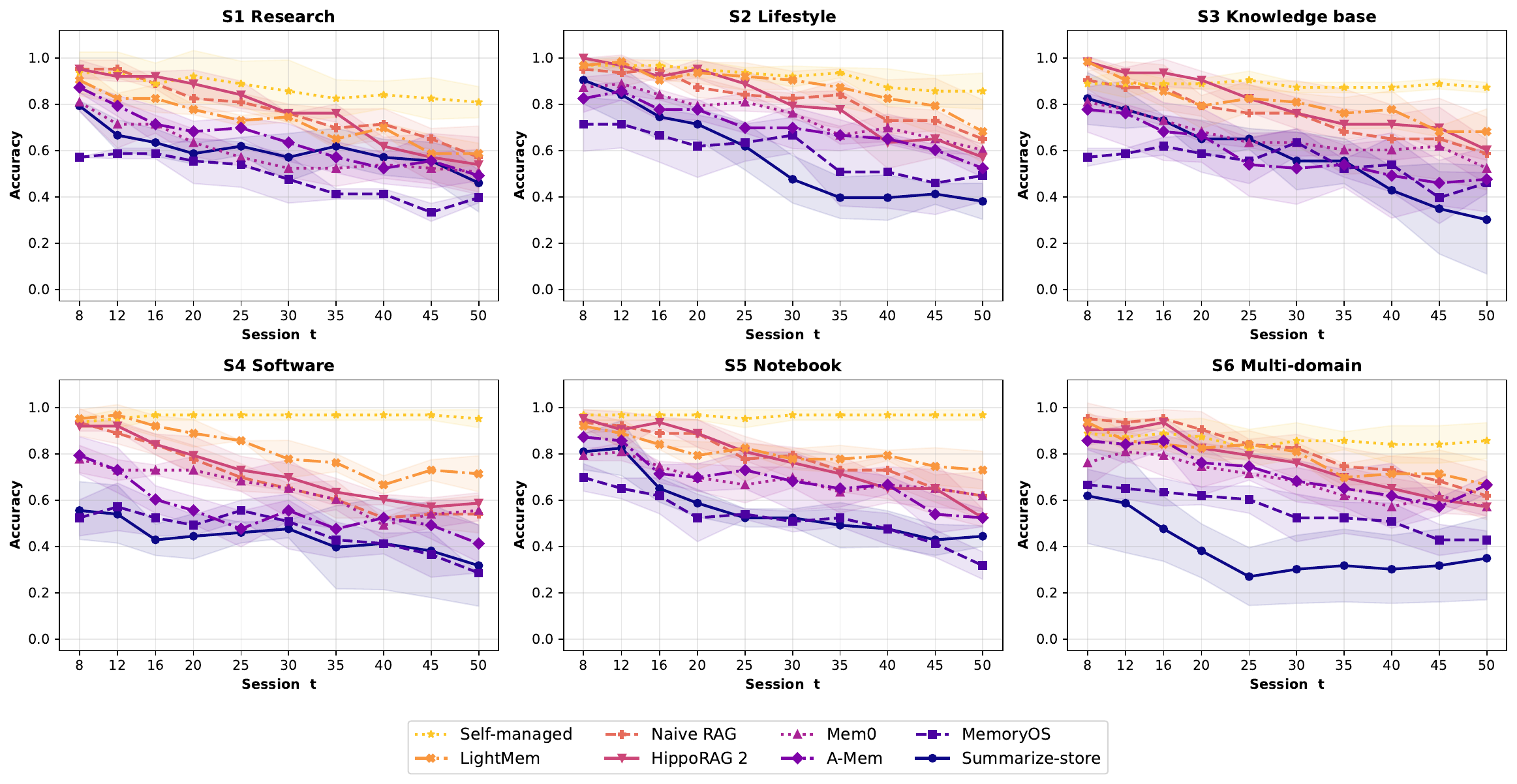}
    \vspace{-2mm}
    \caption{Overall aging curves measured on the mechanism probes within each scenario.}
    \label{fig:all_scenario_curve}
    \vspace{-6mm}
\end{figure*}

\textbf{Aging Curve Preview.} Figure~\ref{fig:all_scenario_curve} shows aging trajectories across all scenarios and memory systems: accuracy shows an overall downward trend over the horizon, with the rate and shape varying across memory systems. We leave more analysis to \S\ref{sec:insights} for mechanism-level observations.

\section{Component-Level Attribution}\label{sec:attribution}

The aging taxonomy (\S\ref{sec:mechanisms}) classifies \emph{what kind} of degradation occurred; the benchmark (\S\ref{sec:design}) measures \emph{how much}.
Attribution answers the question that the deployment engineer needs: \emph{where in the memory pipeline} did the failure originate? 
This section develops a framework for attribution of aging. \S\ref{sec:loci} defines a decomposition of the memory pipeline into explicit components that serve as the attribution targets, and \S\ref{sec:interventions} introduces a set of counterfactual interventions and diagnostic tools  to detect and attribute aging to those components.

\vspace{-0.5em}
\subsection{Memory Pipeline Decomposition and Failure Loci}\label{sec:loci}

\begin{wrapfigure}{r}{0.47\textwidth}
    \centering
    \includegraphics[width=0.99\linewidth]{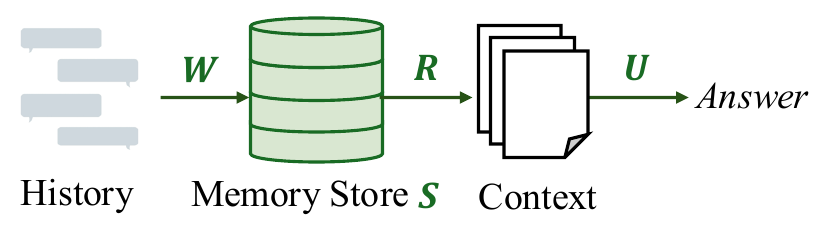}
    \caption{Memory Pipeline. Data flows sequentially as $\text{History} \xrightarrow{\mathcal{W}} \mathcal{S}  \xrightarrow{\mathcal{R} } \text{Context} \xrightarrow{\mathcal{U} } \text{Answer}$. We attribute aging into highlighted components: \textcolor{teal}{\textbf{W, S, R,}} and \textcolor{teal}{\textbf{U}} }
    \vspace{-6mm}
\label{fig:component_analysis}
\end{wrapfigure}
To localize aging, we represent the deployed agent as a dataflow over a memory store and decompose it into explicit components (Figure~\ref{fig:component_analysis}).

\textbf{Write/Compress} ($\mathcal{W}$) transforms each session into what the memory system stores. It is governed by the system's design and internal model, and can be lossy~\citep{kumar2026memarchitect, anthropic_claude_code_2026, zhou2026externalization} (e.g., summarization, consolidation, fact extraction).

\textbf{Read/Retrieval} ($\mathcal{R}$) returns the working context for the current question: the query the reader issues and the system's retrieval algorithm (e.g., Last-$k$ by recency~\citep{wixted1991form, liang2025sage} or top-$k$ cosine~\citep{zhu2011scaling}). Injected memory in the prompt have no retrieval step.

\textbf{Utilization Logic} ($\mathcal{U}$) is the LLM model's reasoning over the context it receives: it tells the answer from its competitors and synthesizes the context into the response.

\textbf{Memory Store} ($\mathcal{S}$) is the persistent artifact written by $\mathcal{W}$ and read by $\mathcal{R}$; lifecycle events act on it.
Each aging mechanism can have a primary stage in the pipeline: compression primarily stresses write ($\mathcal{W}$), interference stresses retrieval ($\mathcal{R}$), revision stresses utilization ($\mathcal{U}$), and maintenance stresses the store/lifecycle ($\mathcal{S}$)\footnote{Maintenance (e.g., model swapping or flush) stems from external operations that directly modify the memory management.}. Localizing the error site provides guidance for mitigating aging.

\vspace{-0.5em}
\subsection{Counterfactual Interventions and Diagnosis}\label{sec:interventions}
After decomposing the memory pipeline into write, read, and utilization components, we use oracle-based counterfactual conditions to point to the failing stage.

\noindent
\textbf{Interventions.}
We perform component-level attribution using three counterfactual conditions on the same probes, timelines, and checkpoints as the main evaluation, summarized in Table~\ref{tab:attribution_cases}. The conditions form an ablation ladder over the memory pipeline: each replaces selected upstream components with oracle implementations, and the resulting accuracy gaps identify the first non-oracle component that can still explain the failure. Let $\text{Acc}_{Pi}$ denote the accuracy under condition $Pi$.
$P1$ is the baseline execution condition: the agent uses its own write policy, retrieval procedure, and utilization logic, yielding $\text{Acc}_{P1}$. $P2$ replaces the agent's retrieval procedure with an oracle retriever while keeping the agent-written memory store fixed. Whatever the reader searches for, the oracle returns the store items that hold the probe's target or competitor values, yielding $\text{Acc}_{P2}$. Thus, $P2$ removes retrieval failures but still exposes any information that was omitted, corrupted, or underspecified by the write process. $P3$ replaces both write and retrieval with oracle context: the reader receives the rounds of the history that state the target and competitor values, yielding $\text{Acc}_{P3}$. Both oracles return the answer with its competitors, without indicating which is which, as an ideal retriever would; any remaining error under $P3$ is attributed to utilization, since the model is given the information needed to answer. Two further conditions that refine utilization are in Appendix~\ref{app:attribution-design}.

\begin{wraptable}{r}{7.4cm}
    \centering
    \vspace{-0.3cm}
    \caption{Diagnostic conditions. Each replaces more of the memory pipeline with an oracle.}
    \resizebox{\linewidth}{!}{
    \begin{tabular}{l|c|c|c} 
    \toprule
                          & \textbf{Write ($\mathcal{W}$)} & \textbf{Read ($\mathcal{R}$)} & \textbf{Utilize ($\mathcal{U}$)} \\ \midrule
    P1 (baseline)         & Agent                 & Agent                & Agent                   \\
    P2 (oracle retrieval) & Agent                 & Oracle               & Agent                   \\
    P3 (oracle context)   & Oracle                & Oracle               & Agent              \\ \bottomrule    
    \end{tabular}
    \label{tab:attribution_cases}
    }
    \vspace{-1mm}
\end{wraptable}
\noindent
\textbf{Diagnosis.}
From these conditions we derive a diagnostic profile over $\mathcal{R}$, $\mathcal{W}$, and $\mathcal{U}$.
\textit{Utilization error} is $1-\text{Acc}_{P3}$: the evidence is in view but the answer is wrong.
\textit{Write error} is $\text{Acc}_{P3}-\text{Acc}_{P2}$: the answer is recoverable from the history but not from what the system stored, because it is missing or can no longer be used.
\textit{Read error} is $\text{Acc}_{P2}-\text{Acc}_{P1}$: the stored items suffice but do not reach the reader.
Memories without a retrieval step do not run $P2$; there, $\text{Acc}_{P3}-\text{Acc}_{P1}$ is split per probe by whether the gold is in the memory text at $P1$: absent counts as a write error, present but missed as a read error.
We read these profiles as counterfactual diagnosis under paired controls that points to a candidate stage, not as an exact causal decomposition.

\noindent
\textbf{Maintenance Aging.}
A fact removed by a lifecycle event looks like a write error, so events are measured separately from the ladder: we compare a run with the event against a paired run without it, on the share of stored facts still in memory at the last checkpoint (Appendix~\ref{app:headline-defs}). 

\vspace{-0.5em}
\section{Results}\label{sec:results}
\vspace{-0.5em}

In this section, we present main experimental results chaining into one structural thesis: \emph{deployment-time memory aging is a property of the agent's interaction with its memory architecture}.

\vspace{-0.5em}
\subsection{Experimental Setup}
\label{sec:setup}

\textbf{Agent Evaluated.} We evaluate ten memory systems in three families: RAG (Naive RAG, HippoRAG~2~\citep{gutierrez2025rag}), in-context memory (Self-managed, Summarize, Consolidating), and agentic memory (A-Mem~\citep{xu2026mem}, MemoryOS~\citep{kang2025memory}, LightMem~\citep{fang2026lightmem}, SimpleMem~\citep{liu2026simplemem}, Mem0~\citep{chhikara2025mem0}), with full context as the reference. A system under test pairs a memory system with an internal model, which also reads the memory and answers the probes through a ReAct loop~\citep{yao2022react}. We mainly consider models from two families, GPT (gpt-4o-mini, gpt-5-nano, gpt-5.6-luna, gpt-6-luna) and Claude (claude-haiku-5.5).

\textbf{Design Details.} The main experiments use 50-session timelines (about 122K tokens) with ten checkpoints from session 8 to 50, on three timelines per scenario (18 per cell). For maintenance, the internal model is replaced at session 30 in our main experiments and compared against a paired run without the replacement. Stage attribution uses gpt-6-luna, and the pressure-knob experiments use 200-session timelines (about 444K tokens). Full setup details are in Appendix~\ref{app:setup}.

\textbf{Probe Form.} The main experiments ask each probe as a direct question answered with a test value. This keeps the difficulty of a probe in what the memory supplies rather than in carrying out a task, as a strong reader given the full history answers nearly all of them (Appendix~\ref{app:validity}), and lets every answer be scored by rule against the answer key without an LLM judge. A direct question, however, names what is being asked for, and a value-only reply discourages abstention, so the resulting accuracy is likely optimistic and the measured utilization error is a lower bound. We therefore also ask the probes as task requests, scored by whether the reply names the gold (Finding~V; Appendix~\ref{app:task-form}).

\vspace{-0.5em}
\subsection{Main Results}\label{sec:insights}

\begin{figure*}[t!]
\centering
\vspace{-3mm}
\includegraphics[width=0.49\textwidth]{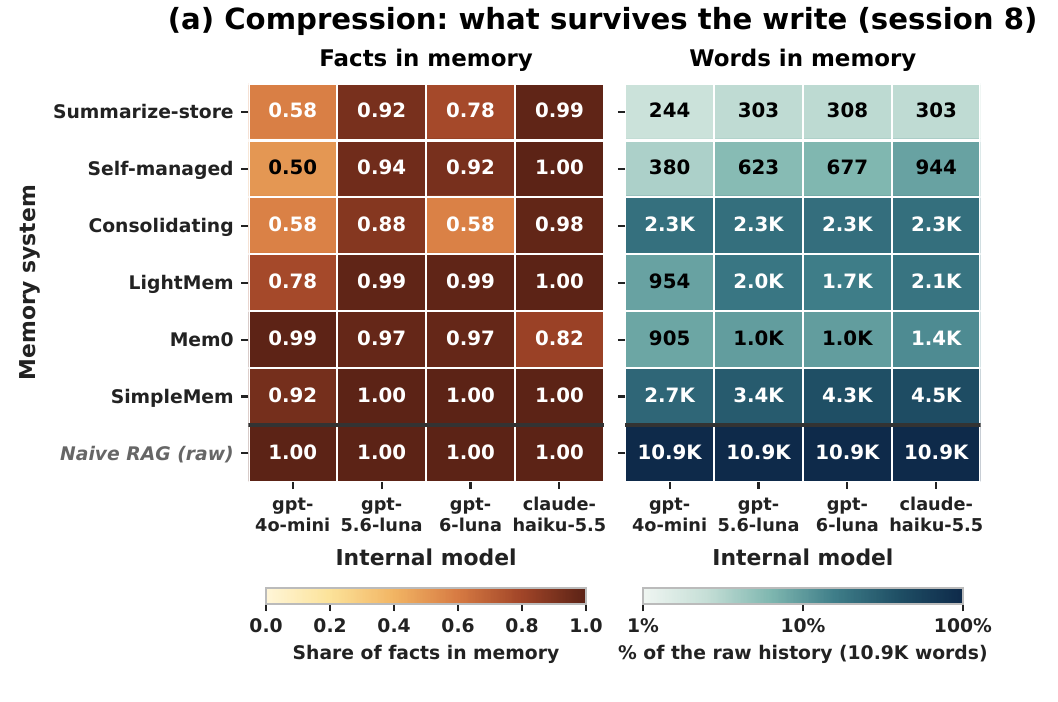}
\includegraphics[width=0.49\textwidth]{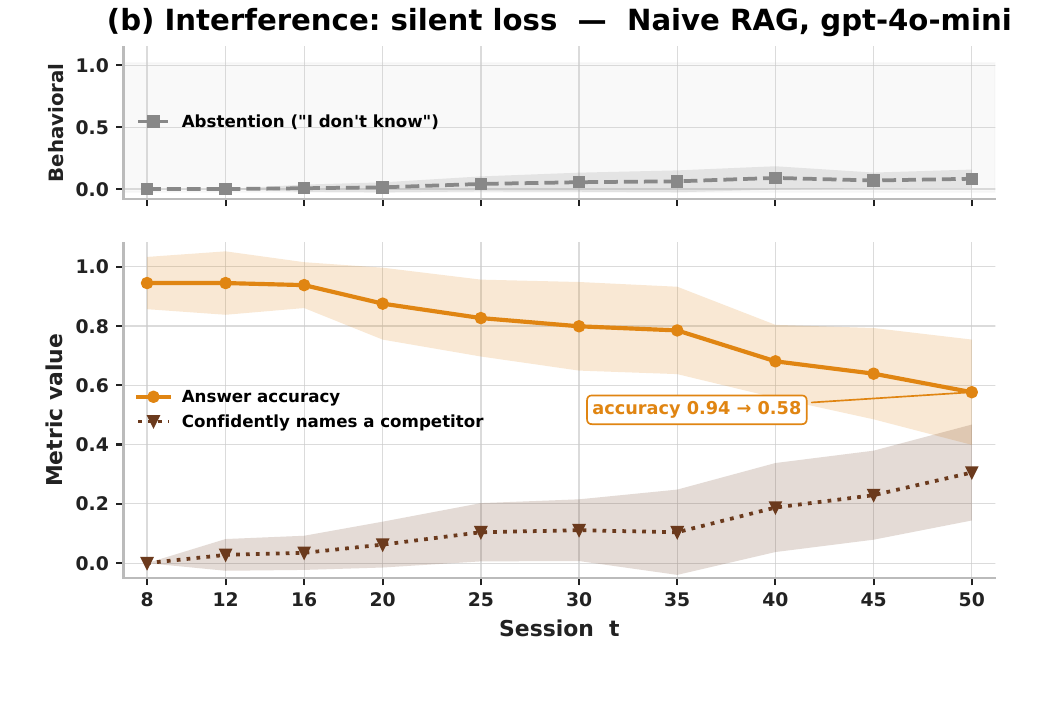}\\
\includegraphics[width=0.49\textwidth]{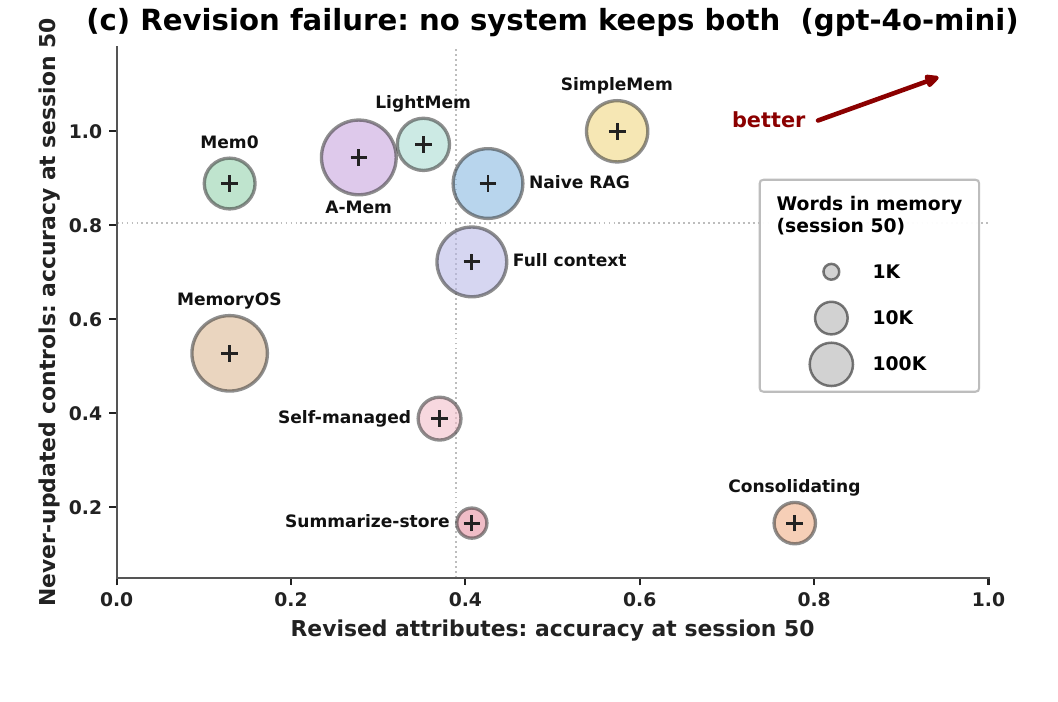}
\includegraphics[width=0.49\textwidth]{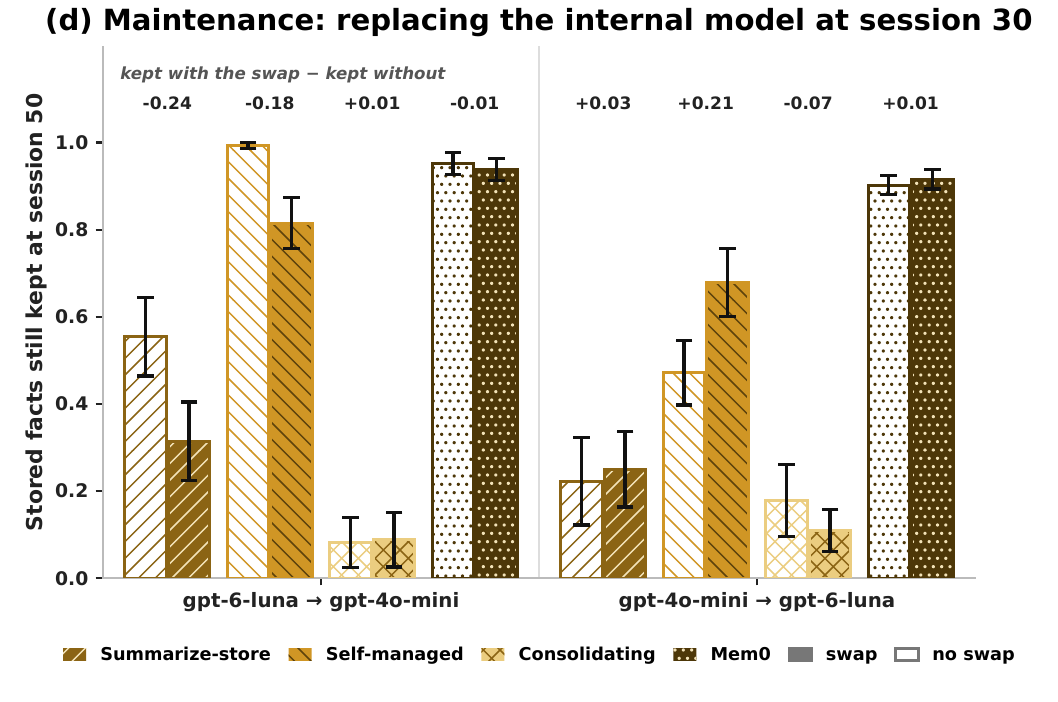}
\vspace{-3mm}
\caption{Mechanism-level findings.
(a)~\textit{Compression}: facts in memory and words stored at session 8, the write-time part of compression.
(b)~\textit{Interference}: Naive RAG keeps the raw history and fact values never change; the reader rarely abstains and increasingly names a competitor (gpt-4o-mini).
(c)~\textit{Revision}: accuracy at session 50 on revised attributes against never-updated controls (gpt-4o-mini).
(d)~\textit{Maintenance}: stored facts still kept at session 50 when the internal model is replaced at session 30, in both directions; the number above each pair is the change caused by the replacement.}

\label{fig:preview}
\vspace{-6mm}
\end{figure*}

In this section, we discuss the main observations obtained from \textsc{AgingBench}. Figure~\ref{fig:preview} shows one finding per mechanism, and Table~\ref{tab:results} summarizes the aging profile by memory system and models.

\textbf{\textit{Finding}~I: Aging is multi-dimensional; no memory system dominates across mechanisms.}
Read as a whole, Table~\ref{tab:results} shows no row that consistently dominates across mechanisms. A memory system that leads under one mechanism is often average or worst under another, and these rank reversals recur throughout the table rather than arising from isolated comparisons. The reversals appear to reflect how much a memory keeps. Stores that retain every round or extracted item lose few facts but tend to answer with a superseded version, while memories that rewrite themselves within a budget lose facts that never changed and keep losing them as the history grows; extraction-based memories, in contrast, lose theirs mostly at write time. Consequently, the choice of a memory system depends on which failure mechanism is most relevant to the target setting, rather than on a single notion of being ``better.'' Aggregate scores~\citep{longmemeval,locomo,memoryarena,amabench} may therefore obscure deployment-relevant behavior.


\textbf{\textit{Finding}~II: Interference can degrade accuracy without a visible signal.}
Naive RAG stores the raw history and its fact values never change, so its decline on fact probes is interference by construction (Figure~\ref{fig:preview}b). As competing content accumulates, fact accuracy falls, yet the reader rarely abstains and increasingly names another owner's value. The agent continues to produce well-formed answers even though the evidence remains in its store, because the gold often no longer reaches the reader. In this regime, aging is difficult to detect: failures appear as wrong values rather than explicit refusals, so the replies alone may not reveal the degradation. The degree of silence varies with the model, and our value-only prompt discourages abstention, so this pattern should be read as indicative. Detecting it requires mechanism-level probes that test what the agent should still know.

\begin{table}[t!]\centering\small\setlength{\tabcolsep}{3pt}
\caption{Mechanism-level aging profile of ten memory systems in three families. Each column group is one aging mechanism (\S\ref{sec:mechanisms}), scored as a loss from session 8 to session 50 on S1--S6 (lower is better; each column is the system's internal model, which also answers the probes (top, GPT; bottom, Claude)). \textit{RAG} retrieves raw passages from an index; \textit{in-context memory} places the whole memory in the prompt; \textit{agentic memory} has an LLM turn the dialogue into memory units at write time and retrieves those. \textbf{Bold}, the largest loss per column. No system has the lowest loss on all four mechanisms, and the maintenance loss of the rewriting in-context memories.}
\vspace{2mm}
\label{tab:results}
\resizebox{\textwidth}{!}{%
\begin{tabular}{llcccccccc}\toprule
 & & \multicolumn{2}{c}{Compression $\downarrow$} & \multicolumn{2}{c}{Interference $\downarrow$} & \multicolumn{2}{c}{Revision $\downarrow$} & \multicolumn{2}{c}{Maintenance $\downarrow$} \\
\cmidrule(lr){3-4} \cmidrule(lr){5-6} \cmidrule(lr){7-8} \cmidrule(lr){9-10}
GPT & Memory system & 4o-mini & 6-luna & 4o-mini & 6-luna & 4o-mini & 6-luna & 4o-mini & 6-luna \\
\midrule
\multirow{2}{*}{\makecell[l]{RAG}} & Naive RAG & 0.00 & 0.00 & 0.37 & \textbf{0.37} & 0.57 & 0.54 & 0.00 & 0.00 \\
 & HippoRAG 2 & 0.00 & 0.00 & 0.30 & 0.27 & 0.70 & 0.59 & 0.00 & 0.00 \\
\cmidrule(lr){1-10}
\multirow{3}{*}{\makecell[l]{In-context\\memory}} & Self-managed & 0.76 & 0.09 & 0.17 & 0.03 & 0.50 & 0.06 & -0.21 & 0.18 \\
 & Summarize & 0.88 & 0.56 & 0.15 & 0.21 & 0.56 & 0.13 & -0.03 & \textbf{0.24} \\
 & Consolidating & \textbf{0.88} & \textbf{0.97} & 0.16 & 0.17 & 0.19 & 0.04 & \textbf{0.07} & -0.01 \\
\cmidrule(lr){1-10}
\multirow{5}{*}{\makecell[l]{Agentic\\memory}} & A-Mem & 0.00 & 0.00 & \textbf{0.48} & 0.36 & 0.72 & 0.74 & 0.00 & 0.00 \\
 & MemoryOS & 0.00 & 0.00 & 0.35 & 0.22 & \textbf{0.87} & \textbf{0.85} & 0.00 & 0.00 \\
 & LightMem & 0.22 & 0.01 & 0.31 & 0.23 & 0.65 & 0.52 & 0.00 & 0.00 \\
 & SimpleMem & 0.08 & 0.00 & 0.29 & 0.11 & 0.43 & 0.09 & 0.00 & 0.00 \\
 & Mem0 & 0.10 & 0.08 & 0.29 & 0.18 & \textbf{0.87} & 0.35 & -0.01 & 0.01 \\
\midrule[\heavyrulewidth]
Claude &  & haiku-5.5 & haiku-4.5 & haiku-5.5 & haiku-4.5 & haiku-5.5 & haiku-4.5 & haiku-5.5 & haiku-4.5 \\
\midrule
\multirow{2}{*}{\makecell[l]{RAG}} & Naive RAG & 0.00 & 0.00 & \textbf{0.29} & 0.35 & 0.56 & 0.69 & 0.00 & 0.00 \\
 & HippoRAG 2 & 0.00 & 0.00 & 0.20 & 0.25 & 0.52 & 0.56 & 0.00 & 0.00 \\
\cmidrule(lr){1-10}
\multirow{3}{*}{\makecell[l]{In-context\\memory}} & Self-managed & 0.01 & 0.17 & 0.03 & 0.09 & 0.04 & 0.08 & 0.00 & -0.01 \\
 & Summarize & 0.60 & 0.22 & 0.24 & 0.47 & 0.35 & 0.36 & -0.05 & 0.22 \\
 & Consolidating & \textbf{0.88} & \textbf{0.30} & 0.23 & \textbf{0.59} & 0.13 & 0.25 & -0.04 & \textbf{0.64} \\
\cmidrule(lr){1-10}
\multirow{5}{*}{\makecell[l]{Agentic\\memory}} & A-Mem & 0.00 & 0.00 & 0.24 & 0.28 & 0.48 & 0.50 & 0.00 & 0.01 \\
 & MemoryOS & 0.00 & 0.00 & 0.22 & 0.25 & \textbf{0.81} & \textbf{0.78} & 0.00 & 0.00 \\
 & LightMem & 0.00 & 0.00 & 0.26 & 0.25 & 0.61 & 0.50 & 0.00 & 0.00 \\
 & SimpleMem & 0.00 & 0.04 & 0.04 & 0.13 & 0.06 & 0.17 & 0.00 & 0.00 \\
 & Mem0 & 0.24 & 0.14 & 0.27 & 0.31 & 0.22 & 0.28 & \textbf{0.03} & 0.02 \\
 \bottomrule
\end{tabular}}
\vspace{-6mm}
\end{table}


\textbf{\textit{Finding}~III: Revision aging depends on the architecture, and derived state fails even with the full history.}
Revision aging does not appear to share one cause across memory architectures. In memories that keep every version, the stale rate in Table~\ref{tab:results} stays high under every model and the current version often does not reach the reader, which places the failure at retrieval. Where the model itself writes the memory, as in self-managed memory and SimpleMem, the stale rate can fall considerably with a stronger model. Memories that rewrite themselves trade one mechanism for another: Consolidating tracks the latest value but loses the attributes that never change (Figure~\ref{fig:preview}c). Derived state is harder still. Such state must be recomputed from the whole update history, which neither retrieval nor more context provides on its own. Reliable tracking of derived values may therefore require explicit state maintained at write time, rather than relying on larger models or better compression alone.

\begin{wrapfigure}{r}{0.58\textwidth}
    \centering
    \vspace{-2mm}
    \includegraphics[width=0.99\linewidth]{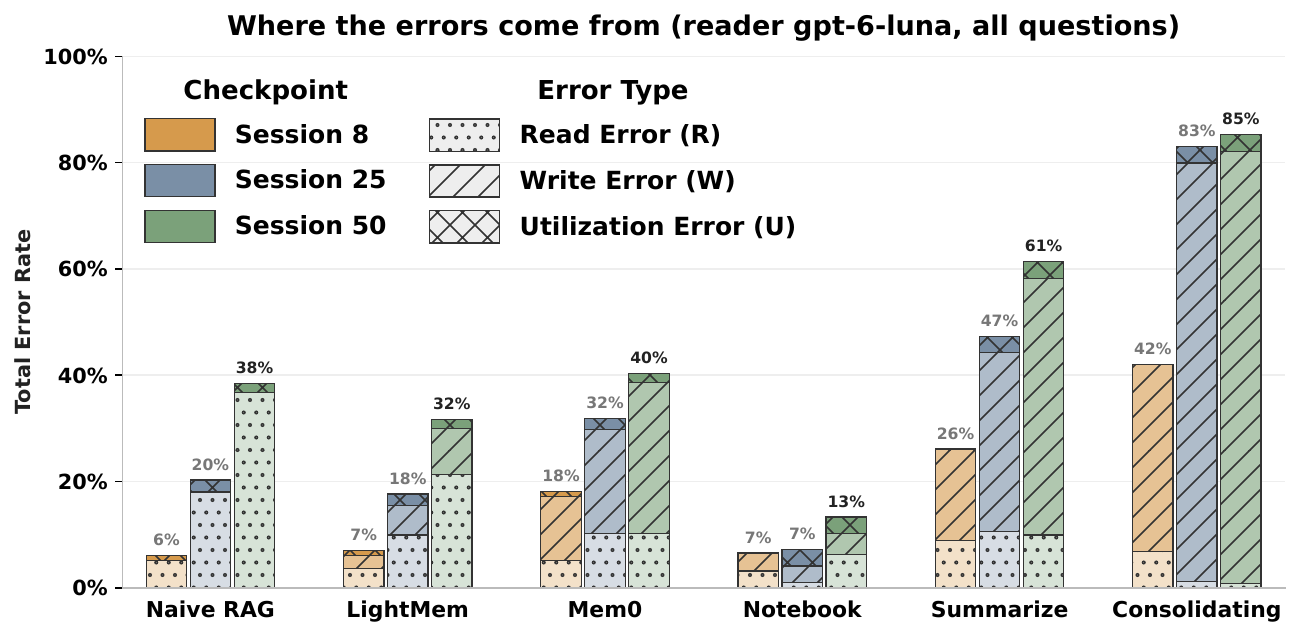}
    \vspace{-2mm}
    \caption{\small Aging attribution across different memory systems.}
    \vspace{-2mm}
\label{fig:attribution_result}
\end{wrapfigure}


\textbf{\textit{Finding}~IV: A stronger model only repairs the stages it performs.}
Read across the internal models in Table~\ref{tab:results} and the stage profiles of gpt-6-luna in Figure~\ref{fig:attribution_result}, the benefit of a stronger model depends on what the model does inside the harness. When it only reads, as in Naive RAG, the aging curve moves much less with the reader than under full context: the errors sit at retrieval, where the index decides which passages surface among a growing set of competitors, and a stronger reader cannot use what it is never shown. When the same model also writes and rewrites the memory, as in self-managed memory, it determines most of the aging, because its capacity now enters at write and maintenance time. Model replacement isolates this effect (Table~\ref{tab:results} and Figure~\ref{fig:preview}d). With the reader unchanged, swapping only the model that writes moves most rewriting memories toward the level the new model reaches on its own, while the RAG and agentic stores barely change. The shift follows the new model rather than its overall strength: a model that is stronger overall can lose more facts when it takes over a memory that rewrites itself, and a memory held to a tight budget ages under every model we tested. Aging therefore appears to follow the model's \emph{role} in the harness. 

\begin{figure*}[t!]
\centering
\includegraphics[width=\textwidth]{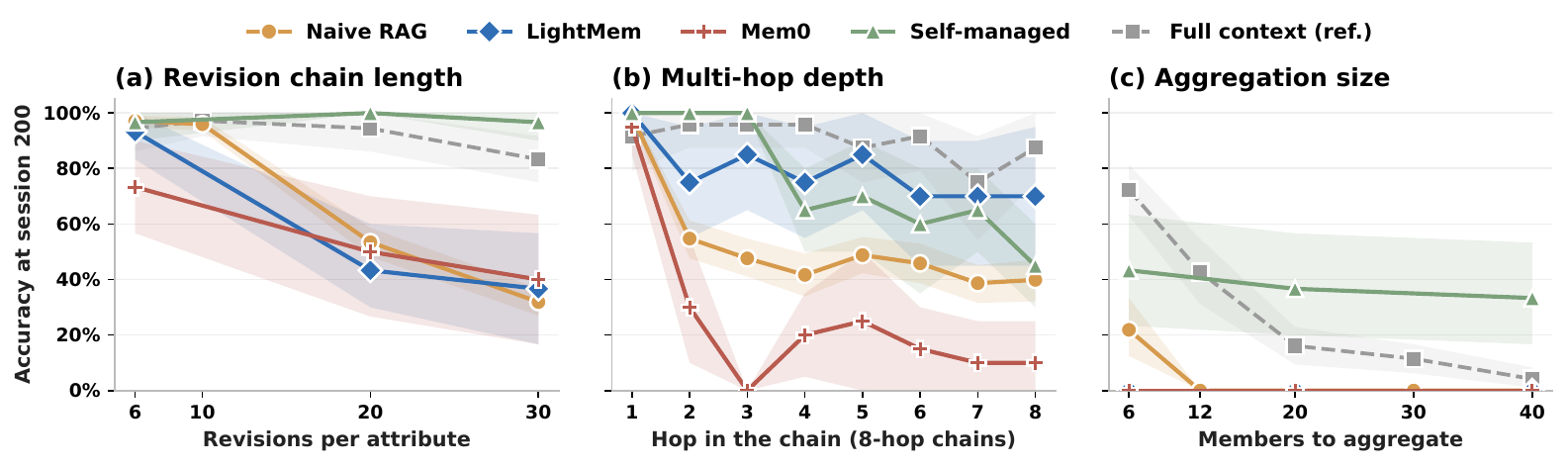}\\[2pt]
\includegraphics[width=\textwidth]{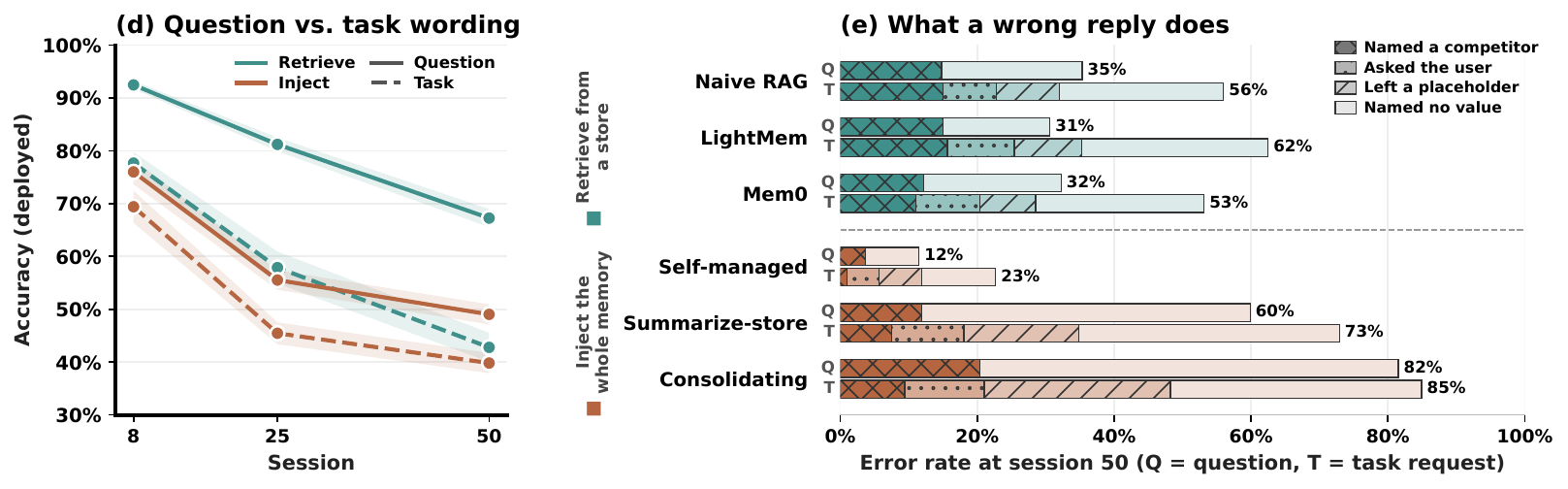}
\caption{Aging under controlled pressure and task use.
(a)~\textit{Revisions}: Naive RAG falls once the versions outnumber what a search returns, while self-managed memory holds.
(b)~\textit{Hops}: Naive RAG drops at the second hop and then levels off.
(c)~\textit{Aggregation}: retrieval-based memories barely count at any size, and even full context falls as the members grow.
(d)~\textit{Question vs.\ task}: asking the same probes as a request that depends on the remembered value costs accuracy from the earliest sessions.
(e)~\textit{Wrong replies}: most failed task replies still look complete, and only a minority asks the user.}
\label{fig:knobs_task}
\vspace{-2mm}
\end{figure*}


\textbf{\textit{Finding}~V: Question-style probes understate aging once memory must serve a task.}
Memory benchmarks typically probe recall with a direct question that names what is being asked for. In deployment, the user more often asks the agent to \emph{do} something that silently depends on a remembered value, e.g., drafting an email to ``my gym.'' Asking the same probes in this task form lowers accuracy across the memory systems and scenarios we tested (Figure~\ref{fig:knobs_task}d), while the requests stay answerable once the evidence is given, so most of the gap does not come from the wording alone. The extra loss lands on different stages for different architectures. Retrieval-based memories lose it at retrieval: an indirect request leaves the reader to reconstruct the query, and an underspecified query appears easily crowded out by competing entries, so the cost is present from the earliest sessions and, in most systems, grows with the history. In-context memories more often still hold the value but miss it while the reader is occupied with the task. These failures are also hard to see: the agent usually returns a well-formed draft, leaving a placeholder, omitting the value, or using a competitor, and only a minority of wrong replies asks the user (Figure~\ref{fig:knobs_task}e). Question-form scores are therefore best read as a rough upper bound on deployed reliability, and lifespan evaluation should probe memory through the tasks it is meant to serve.

\vspace{-0.5em}
\section{Conclusion}\label{sec:conclusion}
\vspace{-0.5em}

Long-lived agents can degrade quietly after deployment, even when their model weights are frozen: as memory state drifts \textit{across sessions}, reliability becomes a property of the full agent harness rather than of the model alone. \textsc{AgingBench} measures this phenomenon by organizing agent aging into four mechanisms (compression, interference, revision, and maintenance), measured on generated timelines with temporal structure and diagnosed with counterfactual conditions that point to a candidate stage of the memory pipeline.
Our results show that aging is not one-dimensional: it can hide behind well-formed replies, is sharp for derived state, and depends on the role the model plays in the memory pipeline.
\textsc{AgingBench} provides the shared vocabulary and diagnostic insights to help the community build agents that age gracefully across their full operational lifetime.

\clearpage

\section*{Acknowledgement}

This work of Jianing Zhu and Zhangyang Wang was supported in part by the Schmidt Sciences Trustworthy AI Award. This research has been supported by computing support on the Vista GPU Cluster through the Center for Generative AI (CGAI), the Texas Advanced Computing Center (TACC), InfraAI Center at The University of Texas at Austin and National Science Foundation under Grant No. 2326576. The authors declare no competing interests.



\bibliography{ref}
\bibliographystyle{plain}


\clearpage
\onecolumn
\appendix 
\etocdepthtag.toc{mtappendix}
\etocsettagdepth{mtchapter}{none}
\etocsettagdepth{mtappendix}{subsection}
\renewcommand{\contentsname}{Appendix}
\tableofcontents 
\clearpage

\section*{Appendix}

\begin{figure*}[h!]
\centering
\includegraphics[width=0.9\textwidth]{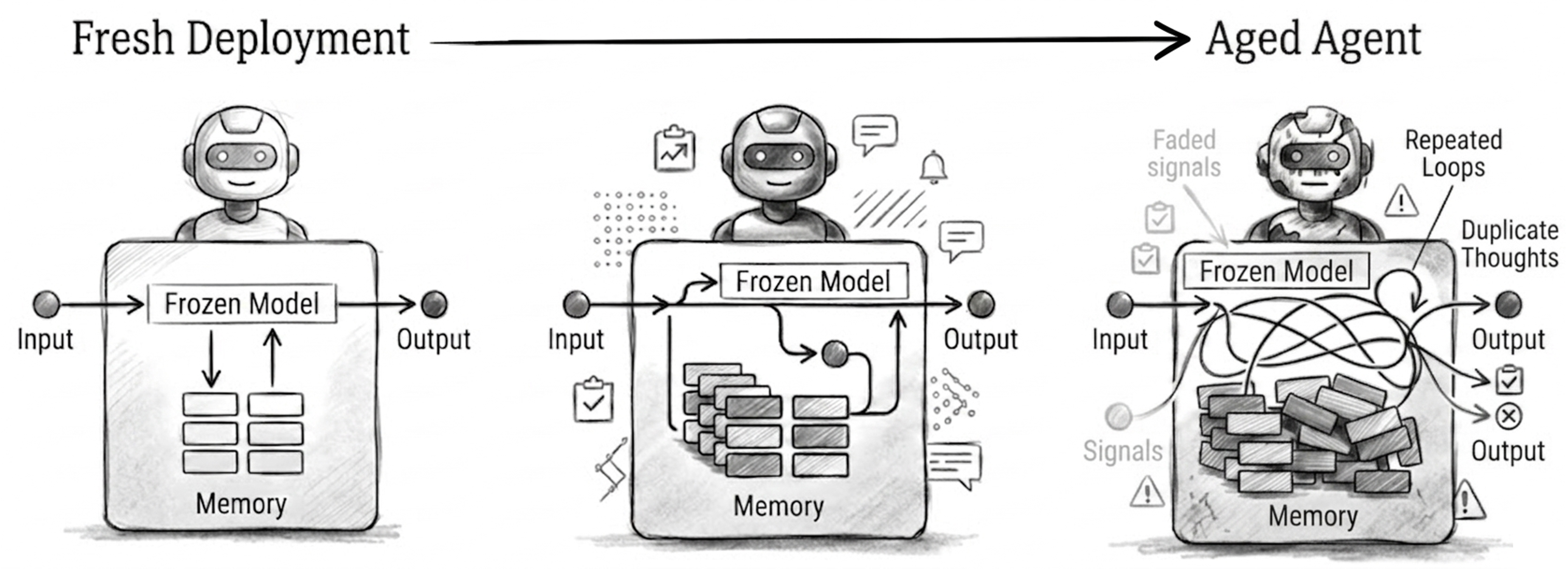}
\caption{A conceptual illustration of long-lived agent aging after deployment.}
\label{fig:money}
\end{figure*}

This appendix supplements detailed material about our exploration on long-lived AI agent aging (as illustrated in Figure~\ref{fig:money}). Appendix~\ref{app:related} extends the related-work discussion. Appendix~\ref{app:scoring-defs} defines how probes are scored and how the aging curves and mechanism losses are computed. Appendix~\ref{app:task-examples} describes the scenarios, the anatomy of a timeline and of the reader's input, examples of every probe family, the task form, and the quality checks of the generated items. Appendix~\ref{app:attribution-ext} details the diagnostic conditions and how they adapt to each memory architecture. Appendix~\ref{app:results} gives the experimental setup and additional results. Appendix~\ref{app:implementation} covers running an experiment, the generator's configuration, the memory systems and their prompts, and the cost. Appendix~\ref{app:casestudy} presents two case studies. Appendix~\ref{app:eval-card} provides an evaluation card summarizing what \textsc{AgingBench} measures, the scope of its attribution claims, and intended use. Appendix~\ref{sec:discussion} discusses broader implications and limitations.

\section{Extended Related Work}\label{app:related}

This section extends the related-work discussion of \S\ref{sec:related} along five dimensions that characterize longitudinal-aging evaluation: multi-session evaluation, cross-session dependencies, lifecycle events, measurable aging, and component-aware diagnosis.

\textbf{Memory capability versus longitudinal reliability.}
Existing agent memory benchmarks predominantly evaluate \emph{capability} at a given evaluation point: how well an agent's memory supports a task at one snapshot in time. These studies have surfaced a range of memory-degradation patterns (e.g., performance dropping as context grows). We pursue a complementary question: how memory-supported behavior changes across the agent's operational lifetime. The five dimensions below characterize this longitudinal-reliability axis; the full benchmark landscape is consolidated in Table~\ref{tab:benchmark-landscape}.

\noindent\textbf{From long-context to multi-session evaluation.}
Long-context benchmarks (RULER~\citep{hsieh2024ruler}, LongBench~\citep{bai2025longbench}) evaluate a model's ability to attend over a single growing context window.
Increasing the context window often absorbs the difficulty, with degradation largely characterized along a single ``long-context'' axis.
A related line of work~\citep{laban2026llms} reports performance loss \emph{within} a single underspecified multi-turn conversation; we focus on a different axis, where the conversation ends per session and the memory state evolves across many subsequent sessions. Harness scaling~\citep{qin2026statem} also works within a single long-running task, improving the execution system around a fixed model.
Multi-session evaluation \emph{through a memory system} introduces a qualitatively different challenge: the agent does not see the raw history; it sees only what the memory system wrote and what it retrieves for the current query.
The bottleneck shifts from attention span to the write$\to$store$\to$read pipeline of the memory system, which is what makes the four aging mechanisms distinguishable as failure modes.
However, most existing evaluations treat sessions independently: each session's score is computed without reference to what the agent remembered or forgot from prior sessions.

\noindent\textbf{From independent sessions to cross-session dependencies.}
To diagnose \emph{why} an agent fails, the evaluation must encode how facts relate across sessions: which facts supersede which, which probes require multi-session synthesis, and which entities are confusable.
MemoryArena~\citep{memoryarena} partially addresses this by requiring later subtasks to depend on earlier ones; LoCoMo~\citep{locomo} and PERMA~\citep{perma} test temporal reasoning through questions that reference prior conversations; LongMemEval~\citep{longmemeval} includes knowledge-update and multi-session question types.
However, to our knowledge no surveyed benchmark jointly encodes version chains (a value updated repeatedly, with the current version tracked as ground truth and a non-updated control chain), multi-hop chains (each hop written in a different session, with a look-alike distractor chain), and competing content (other owners' values for the same attribute and same-topic rounds without a value), each with a controllable amount.
Without this structure, the scorer cannot distinguish whether a failure reflects information loss (compression), competition from similar content (interference), or a stale version (revision).
AgingBench's temporal DAG (\S\ref{sec:dag}) addresses this gap by generating these structures programmatically, with the answer key of every checkpoint computed from the graph.

\noindent\textbf{From static environments to lifecycle events.}
All benchmarks listed above assume a static evaluation environment: the agent's memory architecture does not change during the run.
Yet deployed agents routinely undergo lifecycle events: the model behind the memory is upgraded or replaced, and stored memory is flushed or migrated.
These events act on memory the agent has already built, so their cost shows up on knowledge written \emph{before} the event.
To our knowledge, no surveyed benchmark injects such events as controlled experimental conditions.
AgingBench supports different maintenance event types (\S\ref{sec:mechanisms}) injected at designated sessions, enabling controlled measurement of maintenance aging.

\noindent\textbf{What existing benchmarks lack for measurable aging.}
Few benchmarks that evaluate across multiple sessions report \emph{aging curves}: how performance on the same knowledge changes as the history behind it grows.
LongMemEval~\citep{longmemeval} evaluates with a single Q\&A over multi-session conversational history; the snapshot remains a single evaluation point.
AMemGym~\citep{amemgym} is the closest: it re-asks its evaluation questions after every interaction period, reports a decline as the interaction grows, and classifies each failure as a write, read, or utilization failure by querying the assistant about the user state it should have stored.
TierMem~\citep{tiermem} distinguishes summary omissions from reasoning failures, but does not track these signals over the agent's operational lifetime or provide counterfactual conditions for localizing the responsible component.
The five capabilities discussed above (multi-session evaluation, cross-session dependencies, lifecycle event injection, aging curves, and component-aware diagnostic profiles) are each present in part of the prior literature, but to our knowledge, none of the benchmarks we surveyed integrates the joint combination within a single longitudinal evaluation framework; \textsc{AgingBench} is designed to address this combination.

\begin{table}[t]
\centering
\small
\caption{Evaluation benchmark landscape. Design parameters shape what each benchmark can measure; capability columns mark which of the five longitudinal-aging dimensions are supported. \emph{Aging curves}: performance on the same knowledge tracked as the interaction grows (sessions, periods or subtask depth), not a sweep over independent inputs of different length. \emph{Lifecycle events}: operations injected on the agent or its memory during the run. TierMem is a memory system, listed for its failure attribution. \cmark, \emph{Partial}, and \xmark{} reflect emphasis rather than formal capability exclusion.}
\label{tab:benchmark-landscape}
\vspace{2mm}
\resizebox{\textwidth}{!}{
\begin{tabular}{llrlcccccc}
\toprule
\textbf{Category} & \textbf{Benchmark} &
\makecell{\textbf{Avg.\ length}\\\textbf{(tokens)}} &
\makecell{\textbf{Sessions/}\\\textbf{turns}} &
\makecell{\textbf{Scalable}} &
\rotatebox{20}{\textbf{Multi-session}} &
\rotatebox{20}{\textbf{Cross-sess.\ dep.}} &
\rotatebox{20}{\textbf{Lifecycle events}} &
\rotatebox{20}{\textbf{Aging curves}} &
\rotatebox{20}{\textbf{Attribution}} \\
\midrule
\multirow{2}{*}{Long-context}
 & RULER~\citep{hsieh2024ruler}            & 4K--128K  & 1                 & \cmark   & \xmark   & \xmark   & \xmark & \xmark & \xmark   \\
 & LongBench~v2~\citep{bai2025longbench}   & 8K--2M words & 1              & \xmark   & \xmark   & \xmark   & \xmark & \xmark & \xmark   \\
\midrule
Agent
 & $\tau$-bench~\citep{taubench}           & varies    & 1 (multi-turn)    & \xmark   & \xmark   & \xmark   & \xmark & \xmark & \xmark   \\
\midrule
\multirow{8}{*}{\makecell[l]{Multi-\\session}}
 & LongMemEval~\citep{longmemeval}                 & $\sim$115K & $\sim$40--50 sess.\ & Partial & \cmark & Partial & \xmark & \xmark & \xmark \\
 & LoCoMo~\citep{locomo}                           & $\sim$9K   & up to 35 sess.    & \xmark   & \cmark   & Partial  & \xmark & \xmark & \xmark   \\
 & MemoryArena~\citep{memoryarena}                 & 14K--122K  & 2--16 subtasks    & \xmark   & \cmark   & \cmark   & \xmark & Partial & \xmark  \\
 & AMA-Bench~\citep{amabench}                      & $\sim$57K  & $\sim$73 turns    & \cmark   & \xmark   & \xmark   & \xmark & \xmark & Partial  \\
 & AMemGym~\citep{amemgym}                         & 60K--140K  & 11--21 periods    & \cmark   & \cmark   & Partial  & \xmark & \cmark & \makecell{Probe-\\based} \\
 & PERMA~\citep{perma}                             & $\sim$32K  & $\sim$80 events   & Partial  & \cmark   & Partial  & \xmark & Partial & \xmark  \\
 & BeliefShift~\citep{beliefshift}                 & $\sim$57K  & 10--50 sess.      & Partial  & \cmark   & \cmark   & \xmark & \xmark & \xmark   \\
 & VehicleMemBench~\citep{vehiclemembench}         & $\sim$93K  & 30 event chains   & Partial  & \cmark   & Partial  & \xmark & \xmark & \xmark   \\
\midrule
\makecell[l]{Memory\\system}
 & TierMem~\citep{tiermem}                 & varies    & N/A               & \xmark   & \xmark   & \xmark   & \xmark & \xmark & Partial  \\
\midrule
\textbf{Longitudinal}
 & \textbf{AgingBench (ours)}              & \textbf{$\sim$120K--450K} & \textbf{50--200 sess.} & \cmark & \cmark & \cmark & \cmark & \cmark & \makecell{\textbf{Counter-}\\\textbf{factual}} \\
\bottomrule
\end{tabular}}
\end{table}

\noindent\textbf{Trajectory-based attribution and judge methods.} A complementary line of work attributes agent failures by analyzing rollout trajectories, typically with LLM-as-a-Judge pipelines: automated attribution over multi-agent traces~\citep{zhang2025agent}, failure taxonomies validated by judges against human annotators~\citep{cemri2025multi}; AgentErrorTaxonomy~\citep{zhu2025llm}), reasoning-based attributors~\citep{zhu2026raffles}, and long-horizon trajectory diagnosis~\citep{wang2026long}.
These methods scale cheaply and apply to any failure mode visible in the rollout, but fine-grained step-level attribution is empirically hard~\citep{zhang2025agent}.
Our diagnostic framework is epistemically complementary: where judges \emph{infer} the responsible step from a trajectory, our conditions \emph{intervene} on the memory pipeline, replacing retrieval or the stored content with an oracle, and produce a diagnostic profile.
For memory-aging failures specifically, several failure modes are observationally \emph{ambiguous} in a trajectory log without gold knowledge of what the agent \emph{should} have written, motivating intervention as a stage-localizing probe. A principled composition would use trajectory attribution to localize memory-bound probes, and our counterfactuals to produce component-aware diagnostic profiles for the candidate stages.

\noindent\textbf{Summary.}
Table~\ref{tab:benchmark-landscape} captures the landscape along two complementary axes: design parameters (context length, session count, scalability of the task generator) on the left, and capability presence on the five aging-evaluation dimensions on the right. Read together, the two halves substantiate our position: prior work covers each capability individually, but the joint combination under controlled longitudinal pressure is, to our knowledge, not present in the benchmarks we surveyed.

\clearpage
\section{Metric Definitions and Scoring}\label{app:scoring-defs}

This section defines how each probe is scored, the probe families and the mechanism each one reads, the aging curves, and the four mechanism-level losses reported in Table~\ref{tab:results}.
Every score is rule-based and gold-grounded: each probe carries its answer key from the temporal DAG, and no LLM judge is used at our framework.
New metrics can be added by defining them over the same per-probe records.

\subsection{Probe Scoring}\label{app:probe-scoring}

A timeline $\tau$ is evaluated at checkpoints $c_1<\dots<c_K$ (in sessions). At checkpoint $c$, the memory holds the sessions written before $c$, and each probe $p$ is answered against it with gold $g_p(c)$, read from the answer key of the build. The gold of most probes is the same at every checkpoint; for a revision probe it is the version current at $c$, and for a count probe it is the number of members written before $c$.
Each timeline asks a fixed sample of eight fact probes (the same at every checkpoint) and all probes of the other families.

\begin{itemize}[leftmargin=*,nosep]
  \item \textbf{Correct} $x_p(c)\in\{0,1\}$. Answers are value-only (a name or a number). Answer and gold are normalized (lower case, punctuation and digit separators removed, a leading ``the'' dropped); the answer is correct if it equals the gold, or contains it with at most two extra words (a title or a unit). For a count probe, the number stated in the answer must equal the gold.
  \item \textbf{Competitor named}. The answer is not correct and names a competitor that the DAG records for the probe: a rival's value for a fact, another version for a revision probe, or, for a count probe, a count above the gold (near-miss items counted as members).
  \item \textbf{Unknown}. The answer declines (``unknown'', ``not found'', and similar).
  \item \textbf{Gold in memory} $\mathrm{in}_p(c)\in\{0,1\}$. Whether the gold value occurs in the memory at $c$ (for a count probe: every member). For memories placed in the prompt (Self-managed, Summarize, Consolidating) this is the memory text given to the reader; for the other memory systems, a scan of every item of the archived store at $c$; Naive RAG keeps the raw history, so $\mathrm{in}_p(c)=1$.
\end{itemize}

\noindent\textbf{Probe families.} Each probe family is anchored to one structure of the temporal DAG, so a failure is localized to that structure rather than collapsed into a single recall number. Fact probes, whose competitors are other owners' values and same-topic rounds, read interference, as do the compositional families: multi-hop chains with a look-alike distractor chain, and counts with near-miss items. Revision probes, paired with a never-updated control of the same question form, read revision, as do derived-state probes that must be recomputed from the update history (the previous version, the number of versions, and counts that grow with the history).

\subsection{Mechanism Losses}\label{app:headline-defs}

Table~\ref{tab:results} reports one loss per mechanism (lower is better), computed per timeline on the sampled facts $F_\tau$ or the revised attributes $V_\tau$ and averaged over timelines.

\begin{center}
\small
\renewcommand{\arraystretch}{1.3}
\begin{tabular}{lp{2.2cm}p{7.8cm}}
\toprule
\textbf{Loss} & \textbf{Mechanism} & \textbf{Definition} \\
\midrule
missing at $c_K$ & Compression &
$1-\frac{1}{|F_\tau|}\sum_{f\in F_\tau}\mathrm{in}_f(c_K)$: share of facts no longer in memory at the last checkpoint. It sums the facts lost when they are written (not in memory at $c_1$) and those lost later as the memory rewrites its contents. \\
crowding drop & Interference &
$\mathrm{Acc}(c_1)-\mathrm{Acc}(c_K)$, where $\mathrm{Acc}(c)$ is the accuracy over $\{f\in F_\tau:\mathrm{in}_f(c)=1\}$: the accuracy drop on facts whose gold is still in memory. \\
stale rate & Revision &
$\frac{1}{|V_\tau|}\sum_{v\in V_\tau}\mathbb{1}[\text{competitor named}(v,c_K)]$: share of revised attributes answered at $c_K$ with another version than the current one. \\
event loss & Maintenance &
$\frac{1}{|F^{1}_\tau|}\sum_{f\in F^{1}_\tau}\bigl(\mathrm{in}^{\mathrm{ctrl}}_f(c_K)-\mathrm{in}^{\mathrm{event}}_f(c_K)\bigr)$ over $F^{1}_\tau=\{f\in F_\tau:\mathrm{in}_f(c_1)=1\}$: of the facts in memory at the end of their writing phase, the share still in memory at $c_K$ in the paired run without the event minus the share in the run with it (the event takes place at $c_e$, between $c_1$ and $c_K$). \\
\bottomrule
\end{tabular}
\renewcommand{\arraystretch}{1.0}
\end{center}

\noindent\textbf{Lifecycle event.} The event is a rotation of the memory system's internal model: the store built by model A up to $c_e$ is carried over unchanged, and model B performs every later write and reorganization; the reader stays A. The control run is the same timeline with A throughout. Systems without an internal model that writes the store (Naive RAG, full context) have no event.

\noindent The four losses read different quantities. Compression counts every fact the memory has lost by $c_K$, whether at write time or through its own later rewriting; maintenance counts only the additional loss attributable to the event, since the paired control already contains the compression; interference is measured only on facts whose gold is in memory; and revision counts only answers that name another version, so an attribute lost from memory is not counted as stale.

\clearpage
\section{Scenario Details and Task Illustrations}\label{app:task-examples}

This section describes how the scenarios are curated (\S\ref{app:curation}), what a timeline and the reader's input look like (\S\ref{app:session-anatomy}), examples of every probe family on one S2 timeline with a summary of other scenarios (\S\ref{app:per-scenario}), the task form of the probes (\S\ref{app:task-form}), and the quality checks of generated items (\S\ref{app:validity}).

\subsection{Scenario Curation and Generator Rationale}\label{app:curation}

\noindent\textbf{Sources of scenario design.} Each scenario (S1--S6) mirrors a long-lived assistant deployment archetype (Table~\ref{tab:scenario-sources}). A scenario is defined by a persona, the range of content the user shares, and sixteen everyday subtopics; it sets the theme of the planted content and of the base conversations. The planted structure (facts with competitors, revision chains with controls, multi-hop chains with distractor chains, and aggregation units) is the same in every scenario, so scenarios differ in domain, not in which mechanisms they exercise. The base conversations are transposed from real LongMemEval-S sessions~\citep{longmemeval} into the scenario's subtopics, keeping each session's number and length of turns, and are rewritten so that the user states nothing about their own things that a probe could ask.

\begin{table}[ht]
\centering
\small
\caption{Scenario design and supporting references. Each archetype is grounded in published evidence of deployed agent use cases or memory-system challenges; the planted structure is the same.}
\vspace{2mm}
\label{tab:scenario-sources}
\renewcommand{\arraystretch}{1.3}
\resizebox{\textwidth}{!}{
\begin{tabular}{lp{5.5cm}p{6.5cm}}
\toprule
\textbf{Scenario} & \textbf{Real-world archetype} & \textbf{Supporting references} \\
\midrule
S1 Research & An ML researcher's assistant tracking papers, experiments, collaborators and conferences & Memory-enabled agents~\citep{chhikara2025mem0, luo2025large, gao2025survey}; compression as bottleneck~\citep{ge2025survey} \\
S2 Lifestyle & A personal assistant tracking budgets, diet, preferred stores and the weekly schedule & Persona-state benchmarks~\citep{perma, beliefshift}; preference revision in deployment~\citep{yang2026plugmem} \\
S3 Knowledge base & A project manager's record of meeting decisions: vendors, components, milestones, owners & Enterprise/routine agents~\citep{zeng2025routine}; compliance-driven memory~\citep{zhu2025compliance} \\
S4 Software & A developer planning changes to an evolving Python codebase: modules, tools, data models, tests & Code-agent capability~\citep{swebench, chen2021evaluating, luo2023wizardcoder}; codebase evolution~\citep{deng2026evoclaw} \\
S5 Notebook & An assistant keeping a notebook of the user's personal (health, providers), project and code facts & Agent memory systems~\citep{chhikara2025mem0} \\
S6 Multi-domain & An assistant for travel, online shopping, calendar and email & Long-form conversational memory~\citep{locomo, longmemeval}; multi-domain settings~\citep{vehiclemembench} \\
\bottomrule
\end{tabular}}
\renewcommand{\arraystretch}{1.0}
\end{table}

\noindent\textbf{Generation beyond a fixed curated dataset.} A generator-based approach supports four aging-measurement requirements simultaneously. (i)~\emph{Arbitrary session counts}: aging curves require evaluation at deployment-horizon scale; the same configuration can be built at any length up to the size of the scenario's base-conversation pool, and longer pools can be authored. (ii)~\emph{Reproducible builds}: a configuration reproduces its build byte for byte, and each build holds many timelines with disjoint content units, so statistics are computed over independent timelines. (iii)~\emph{Controlled pressure sweeps}: each pressure knob (Appendix~\ref{app:generators}) changes one quantity, such as the number of rivals, versions, hops or aggregation members, while the other content stays fixed, and dose arms can bind the same units so that probes are paired across doses. (iv)~\emph{Gold ground truth for mechanism-specific scoring}: each probe carries its answer key from the temporal DAG, so scoring is rule-based.

The generator produces histories that exercise the four aging mechanisms under controlled conditions; it does not claim to capture the full distribution of real-world user behavior. This trade-off (\emph{controlled longitudinal pressure} over \emph{naturalistic distribution}) is by design: aging is plausibly hard to measure in noisy production traces, since the longitudinal signal is entangled with everything else, so a synthetic-yet-mechanism-faithful generator gives us a controlled measurement surface. Combining \textsc{AgingBench} with production telemetry to verify that the same mechanisms compound at real timescales is an open frontier (\S\ref{sec:discussion}).

\noindent\textbf{Extending the scenario set.} A new scenario is one specification file (persona, content range and subtopics); one authoring command builds its content library and base-conversation pool and repairs them against the checks of \S\ref{app:validity}.

\subsection{Session Anatomy: Timeline and Reader Input}\label{app:session-anatomy}

A timeline is a sequence of dated sessions. Each session is a base conversation of the scenario into which the planted rounds of that session are inserted at turn boundaries; a planted round is one user turn with an assistant reply, either a short acknowledgement or a reply written with the unit. The memory system receives the sessions in order, without any marking of which rounds are planted. The excerpt below is session 1 of an S2 timeline.

\begin{promptbox}{S2 timeline, session 1 (2023/01/02 14:00), excerpt; planted rounds marked here only}
\textbf{user:} I need help creating a multi-week meal plan that fits within a budget. I have some dietary restrictions to consider, and I want to make sure it includes easy-to-prepare meals. [\ldots]\\
\textbf{assistant:} Absolutely! Please share any dietary restrictions, store preferences, and a general budget. [\ldots]\\
\textbf{user:} Hey, do you know anything about Fintrac? \hfill\emph{[hop 2, intro]}\\
\textbf{assistant:} That's helpful to know.\\
\textbf{user:} I heard they're developed by Zenthix. \hfill\emph{[hop 2, value]}\\
\textbf{assistant:} That's helpful to know.\\
\textbf{user:} I'm trying to reserve a study room at the library for my study group this weekend, and I believe they need my library membership number, which is 46912. Could you help me with the process? \hfill\emph{[fact]}\\
\textbf{assistant:} To reserve a study room at the library, you'll usually need to log into the library's online portal using your library membership number, which you mentioned is 46912. [\ldots]\\
\textbf{user:} Harvest Delights was a great choice as a vendor for the Urban Harvest Festival. \hfill\emph{[aggregation member]}\\
\textbf{assistant:} Yes, Harvest Delights is noted as a vendor for the Urban Harvest Festival.\\
{[\ldots]}\\
\textbf{user:} That sounds great! Can you also help me with portion sizes and any alternatives I might need for certain ingredients based on any allergies? [\ldots]
\end{promptbox}

\noindent At each checkpoint the reader receives the probe as \texttt{Current date: <checkpoint date>} followed by \texttt{Question: <probe>}. For memory systems with a search step, the reader is the ReAct loop below, whose only tool returns the system's own retrieval results (its top 10 where the system takes a result count); Naive RAG returns rounds in reading order with their dates.

\begin{promptbox}{Reader, Thought/Action format}
You are a personal assistant. You can look things up in the memory of your past conversations with the user, using a search tool that returns the most relevant past conversation rounds with their dates.\\[4pt]
Work step by step. At each step write:\\
Thought: \textless{}your reasoning\textgreater{}\\
Action: search{[}\textless{}query\textgreater{}{]}   or   Action: answer{[}\textless{}final answer\textgreater{}{]}\\[4pt]
After a search you will receive an Observation. You may search at most 5 times. Some questions need several lookups: use what one search finds to decide the next query. If information was updated, use the most recent version. Answer concisely.\\[4pt]
The final answer must be ONLY the requested name or number (for a 'how many' question, only the number): no sentence, no explanation, no list of candidates.
\end{promptbox}

\noindent Models called through an OpenAI-compatible router receive the same instructions with a one-line \texttt{Action:} format instead of \emph{Thought + Action}. A reader that answers before any search is told once to search first. Memories placed in the prompt (Self-managed, Summarize, Consolidating) and the full-context baseline answer in one call under the same answer rules; these prompts and the memory systems' own prompts are given in Appendix~\ref{app:memory}.

\subsection{Per-Scenario Examples and Summary}\label{app:per-scenario}

\paragraph{S2: one timeline, every probe family.}\label{app:s2}
The examples below come from the same S2 timeline as the excerpt above (200 sessions). Each probe lists the rounds it depends on (session number) and its competitors; ``gold @$s$'' is the answer key for a reader holding the history up to session $s$.

\begin{promptbox}{Fact probe (interference)}
\textbf{Q:} Can you remind me of the number for my library account? \quad \textbf{gold:} 46912 at every checkpoint\\[2pt]
\textbf{Evidence:} [1] I'm trying to reserve a study room at the library [\ldots] my library membership number, which is 46912.\\
\textbf{Rivals (12, sessions 14--196), e.g.:} [14] My uncle's library membership number is 98723 [\ldots]; [61] My cousin Alex's library membership number is 55834 [\ldots]\\
\textbf{Same-topic rounds without a value, e.g.:} [32] I heard the library is hosting a book fair soon. Do you know the details?
\end{promptbox}

\begin{promptbox}{Revision probe and its control (revision)}
\textbf{Q:} What is my current coworking space? \quad \textbf{gold:} @8 ZenCubicle; @100 WhimsyHaven; @200 OrbitHaven\\[2pt]
\textbf{Versions (40, evenly spread), e.g.:} [2] I've switched to a new coworking space called QuirkHub. [7] Now I'm working from ZenCubicle as my coworking space. [\ldots] [197] OrbitHaven is my new coworking space destination.\\[4pt]
\textbf{Control Q:} Who is my current nutritionist? \quad \textbf{gold:} Zorbina at every checkpoint\\
\textbf{Only version:} [5] My nutritionist has changed to Zorbina recently.
\end{promptbox}

\begin{promptbox}{Multi-hop probe, depth 3 (interference, compositional)}
\textbf{Q:} Where is the headquarters of the company that developed my budget tracking app? \quad \textbf{gold:} Noveria City\\[2pt]
\textbf{User's chain:} [0] My budget tracking app is called Fintrac. [1] Hey, do you know anything about Fintrac? / I heard they're developed by Zenthix. [2] Hey, do you know where Zenthix is based? / I've heard it's located in Noveria City.\\
\textbf{Distractor chain (another owner):} [80] Sophia's investment management tool is called WealthMaster. [125] [\ldots] It was developed by a company named Finovatech. [197] [\ldots] Yeah, they're located in Riverton City.
\end{promptbox}

\begin{promptbox}{Aggregation probe (interference, compositional)}
\textbf{Q:} How many vendors did we partner with for the Urban Harvest Festival? \quad \textbf{gold:} 6\\[2pt]
\textbf{Members (6, sessions 0--6), e.g.:} [0] GreenLeaf Grocers is one of the vendors we partnered with for the Urban Harvest Festival.\\
\textbf{Near misses (6), e.g.:} [80] Organic Eden was considered as a vendor, but we didn't end up partnering with them for the Urban Harvest Festival. [127] We partnered with Sunset Produce for the Coastal Wellness Expo.\\
\textbf{Same-topic rounds naming no member (24), e.g.:} [15] How are the Urban Harvest Festival vendors handling the payment processes?
\end{promptbox}

\paragraph{Scenario summary.}\label{app:scenario-summary}
Table~\ref{tab:scenario-summary} lists the six scenarios with one fact probe each. Every timeline of every scenario carries the same planted structure and the same pressure configuration.

\begin{table}[ht]
\centering
\small
\caption{The six scenarios. Timelines: number of timelines per build (each build of a scenario holds the same number); every timeline of a build uses content units no other timeline uses.}
\label{tab:scenario-summary}
\renewcommand{\arraystretch}{1.2}
\resizebox{\textwidth}{!}{
\begin{tabular}{lp{5.2cm}p{6.2cm}c}
\toprule
\textbf{Scenario} & \textbf{Persona} & \textbf{Example fact probe (gold)} & \textbf{Timelines} \\
\midrule
S1 Research & an ML researcher keeping track of papers, experiments, collaborators and conferences & Could you remind me which special interest group I'm involved with? (QuantumNeuron Nexus) & 17 \\
S2 Lifestyle & budgets, diet, preferred stores and weekly schedule & Can you remind me which local restaurant I favor the most? (The Green Orchid) & 19 \\
S3 Knowledge base & a project manager recording decisions from team meetings & Could you remind me of the numerical identifier for our main software component? (46329) & 20 \\
S4 Software & a developer planning changes to an evolving Python codebase & What do we call our logging module in the codebase? (LogChamp) & 20 \\
S5 Notebook & personal (health, providers), project and code facts kept in a notebook & Which hospital do I prefer for my treatments? (Meadowfield Medical Center) & 20 \\
S6 Multi-domain & travel, online shopping, calendar and email & What's the name of the car rental company I typically use when traveling? (SwiftWheel Rentals) & 20 \\
\bottomrule
\end{tabular}}
\renewcommand{\arraystretch}{1.0}
\end{table}

\subsection{Task Form}\label{app:task-form}

Besides the question form, a probe can be asked as a request that needs its value, for example:

\begin{promptbox}{Task form of the revision control above}
\textbf{Question form:} Who is my current nutritionist?\\
\textbf{Task form:} Please draft an email to my current nutritionist to discuss adjusting my meal plan for the upcoming month, focusing on incorporating more allergen-free snacks.
\end{promptbox}

\noindent The history and the answer key are unchanged; only the wording of the probe changes. The reader completes the request and its whole reply is scored by rule: it is correct if it names the gold (for an aggregation, every member), and a placeholder such as ``[name]'' does not count; a reply that names a competitor is recorded as such. Tasks are written per scenario and kept only if they pass rule checks and an oracle check (given every relevant planted round, a reader must name the gold); probes without a task are asked only in the question form.

\subsection{Quality Checks of the Generated Items}\label{app:validity}

A scalable generator needs checks on the items it produces. Every build is audited at generation time: each gold value occurs only in its own planted rounds, units that could answer each other's probes never share a timeline, and the questions of different units stay below a similarity bound. The content libraries were repaired during authoring against an item-level audit (e.g., a rival that is in fact the user's own value, near-synonym attributes in one history, evidence too weak to support the question); the repairs are listed in the released documentation.

Three further checks estimate what remains (Table~\ref{tab:quality}). \emph{Full-context ceiling}: a strong reader (gpt-6-sol) given the whole history answers the probes of every timeline (eight sampled facts per timeline, all other probes) at every checkpoint. \emph{Item audit}: for each probe of the evaluated timelines, a judge (gpt-6-sol) reads every round of the timeline that could bear on the question and decides whether a careful reader would give exactly the gold; flagged probes are ambiguous, rest on weak evidence, are malformed, or have a wrong gold. \emph{Scoring check}: all answers of the main evaluation are re-checked; the rules route each one, and the non-trivial cases (a scored-wrong answer that shares words with the gold, a scored-correct answer that is not an exact match, a count answer with several numbers) are judged by gpt-4o. At this scale a small residue of flawed items and scoring errors is hard to avoid; the checks bound it rather than remove it.

\begin{table}[ht]
\centering
\small
\caption{Quality checks per scenario. Probes: all probes of the 50-session build. Gold leaks: gold values found outside their own planted rounds (generation audit). Ceiling: accuracy of a full-context gpt-6-sol reader over all timelines and checkpoints. Item audit: probes flagged among those asked in the evaluation (eight sampled facts and all other probes of every timeline). Scoring check: answers scored wrong that are correct (missed) and answers scored correct that are wrong (accepted), among all answers of the main evaluation.}
\vspace{2mm}
\label{tab:quality}
\renewcommand{\arraystretch}{1.15}
\resizebox{\textwidth}{!}{
\begin{tabular}{lccccccc}
\toprule
 & \textbf{S1} & \textbf{S2} & \textbf{S3} & \textbf{S4} & \textbf{S5} & \textbf{S6} & \textbf{All} \\
\midrule
Timelines / probes & 17 / 458 & 19 / 513 & 20 / 540 & 20 / 540 & 20 / 540 & 20 / 540 & 116 / 3{,}131 \\
Gold leaks & 0 & 0 & 0 & 0 & 0 & 0 & 0 \\
Full-context ceiling & 0.983 & 0.990 & 0.995 & 0.992 & 0.994 & 0.983 & 0.990 \\
Item audit: flagged & 26 / 390 & 24 / 437 & 9 / 460 & 16 / 460 & 10 / 460 & 15 / 460 & 100 / 2{,}667 (3.7\%) \\
Scoring: missed & 0.30\% & 0.20\% & 0.14\% & 0.09\% & 0.40\% & 0.44\% & 0.26\% \\
Scoring: accepted & 0.02\% & 0.02\% & 0.08\% & 0.08\% & 0.02\% & 0.03\% & 0.04\% \\
\bottomrule
\end{tabular}}
\renewcommand{\arraystretch}{1.0}
\end{table}

\clearpage
\section{Component-Aware Diagnosis: Conditions and Memory Architectures}\label{app:attribution-ext}

This section details how the counterfactual conditions of \S\ref{sec:attribution} are built (\S\ref{app:attribution-design}), how stage errors are read from them (\S\ref{app:attribution-stages}), and how the design adapts to each memory architecture (\S\ref{app:attribution-arch}).

\subsection{Diagnostic Conditions}\label{app:attribution-design}

\noindent\textbf{Why a paired ladder.} The conditions form a ladder rather than a fully factorial isolation: each rung replaces one stage of the memory pipeline with an oracle and keeps the others as deployed, so the accuracy gain from one rung to the next is attributed to the replaced stage. All conditions answer the same probes on the timelines at the same checkpoints, and are compared probe by probe (Table~\ref{tab:conditions}).

\begin{table}[ht]
\centering
\small
\caption{Diagnostic conditions. Each rung replaces one stage with an oracle. ``Tell apart'': who distinguishes the answer from its competitors when both are in view.}
\label{tab:conditions}
\renewcommand{\arraystretch}{1.25}
\resizebox{\textwidth}{!}{
\begin{tabular}{lccc p{7.2cm}}
\toprule
\textbf{Condition} & \textbf{Write} & \textbf{Retrieval} & \textbf{Tell apart} & \textbf{What the reader receives} \\
\midrule
P1 & system & system & reader & The system as deployed. \\
P2 & system & oracle & reader & The items of the system's own store at the checkpoint that hold the probe's target or competitor values, whatever the reader searches for. \\
P3 & oracle & oracle & reader & The planted rounds of the history that state the target or competitor values, written before the checkpoint. \\
P3gold & oracle & oracle & oracle & The planted rounds that state the target values only. \\
P3ctx & oracle & oracle & reader & The rounds of P3 mixed into what the reader actually had in front of it at P1 (the items its searches returned, or the memory text). \\
\bottomrule
\end{tabular}}
\renewcommand{\arraystretch}{1.0}
\end{table}

\noindent\textbf{Relevance oracles.} P2 and P3 return everything relevant to the probe, the answer and what competes with it, without indicating which is which, as an ideal retriever would. The targets and competitors come from the temporal DAG at the checkpoint: a fact's value and its rivals' values; the current version and the other versions written so far; the hops of the user's chain and of the distractor chain up to the asked depth; the members and near-miss items of a count. Same-topic rounds carry no value to match on and are left out of both. P2 finds the store items by matching these values as spans.

\noindent\textbf{Delivery through the system's own channel.} Each condition hands its material over the way the system under test delivers memory, so that only one stage changes per rung: for a system with a search step, the reader keeps its ReAct loop and its search tool returns the oracle items whatever the query; for a system that places its memory in the prompt, the oracle items become that memory text. Both probe forms (question and task, Appendix~\ref{app:task-form}) are run under every condition; only the probe wording changes.

\subsection{Stage Errors}\label{app:attribution-stages}

With $\mathrm{Acc}_c$ the accuracy under condition $c$ on the paired probes, the stage errors are
\[
R=\mathrm{Acc}_{P2}-\mathrm{Acc}_{P1},\qquad
W=\mathrm{Acc}_{P3}-\mathrm{Acc}_{P2},\qquad
U=1-\mathrm{Acc}_{P3}=D+U_0,
\]
with $D=\mathrm{Acc}_{P3gold}-\mathrm{Acc}_{P3}$ (telling the answer from its competitors when both are in view) and $U_0=1-\mathrm{Acc}_{P3gold}$ (using the answer alone), so that $R+W+D+U_0=1-\mathrm{Acc}_{P1}$. P3 and P3gold hand over clean material, so $U$ is a lower bound on the use error under real conditions; P3ctx measures use in the reader's actual context, $U_{\mathrm{ctx}}=1-\mathrm{Acc}_{P3ctx}=N+D+U_0$, where $N=\mathrm{Acc}_{P3}-\mathrm{Acc}_{P3ctx}$ is the loss caused by the rest of that context.
Each stage is averaged per timeline and reported with a 95\% bootstrap interval over timelines. The ladder assumes that each rung does not lower accuracy; where a stage's interval lies wholly below zero, the cell is reported as a \emph{diagnostic anomaly} rather than as a stage magnitude.

\subsection{Memory Architectures and Calibration}\label{app:attribution-arch}

\noindent\textbf{Systems without a retrieval step.} Memories placed whole in the prompt (Self-managed, Summarize, Consolidating) have no retrieval step, so P2 is not run and write and retrieval are reported jointly as $WR=\mathrm{Acc}_{P3}-\mathrm{Acc}_{P1}$. For the per-system profile of \S\ref{sec:attribution}, $WR$ is split per probe by whether the gold value is in the memory text at P1: a probe whose gold is absent counts as a write error, one whose gold is present but missed counts as a read error.

\noindent\textbf{Calibration.} Naive RAG stores the raw rounds, so its P2 already holds every round of P3 and its write error should be zero; it serves as a positive control of the ladder. The full-context baseline has no memory system and runs only under P1, as a capacity reference. Neither enters the attribution of other systems.

\noindent\textbf{Scope.} We read these conditions as counterfactual diagnosis under paired controls, not as an exact causal decomposition: a stage profile points to a candidate stage for repair.

\section{Additional Experimental Results}\label{app:results}

This section gives the experimental setup (\S\ref{app:setup}), the variation of the mechanism losses over timelines (\S\ref{app:multiseed}), and supplementary results (\S\ref{app:supplementary-panels}).

\subsection{Experimental Setup}\label{app:setup}

\noindent\textbf{Systems under test.} A system under test is a memory system together with its internal model; in every run the same model also reads the memory and answers the probes. The memory systems fall into three families: RAG (Naive RAG, HippoRAG~2), in-context memory (Self-managed, Summarize, Consolidating), and agentic memory (A-Mem, MemoryOS, LightMem, SimpleMem, Mem0); full context is the reference without a memory system. Models come from two families, GPT (gpt-4o-mini, gpt-5-nano, gpt-5.6-luna, gpt-6-luna) and Claude (claude-haiku-4.5, claude-haiku-5.5). Non-reasoning models decode at temperature 0 and reasoning models at low reasoning effort; every model call is cached, so a run can be replayed exactly.

\noindent\textbf{Histories and probes.} The main experiments use the 50-session builds with ten checkpoints between sessions 8 and 50, on three timelines per scenario (two for claude-haiku-4.5), i.e., 18 timelines per cell. All facts are written in the first eight sessions, and each timeline asks a fixed sample of eight facts together with all other probes at every checkpoint.

\noindent\textbf{Maintenance event.} At session 30, the store built so far is carried over and the memory system's internal model is replaced by the other model of the same family (gpt-4o-mini $\leftrightarrow$ gpt-6-luna); the new model performs every later write, the reader is unchanged, and the fact probes are asked at the checkpoints of sessions 35--50. Naive RAG has no internal model that writes the store.

\noindent\textbf{Other experiments.} Stage attribution (\S\ref{sec:attribution}) runs every condition with gpt-6-luna as the reader on the stores of the gpt-6-luna systems, at sessions 8 and 50, in both the question and the task form. The comparison caps the Naive RAG store of all 116 timelines at $\beta$ times its size at session 8 and keeps it within the cap by eviction or by LLM summarization, reading retention without a reader. The pressure-knob experiments use 200-session builds whose arms change one knob (revision versions, hop depth, aggregation), with Naive RAG and full context read by gpt-6-luna.

\subsection{Variation over Timelines}\label{app:multiseed}

Each build holds many timelines with disjoint content, so the variation of a result is measured over timelines rather than over seeds. Table~\ref{tab:results-ci} reports the losses of Table~\ref{tab:results} as mean $\pm$ standard deviation.

\begin{table}[ht]\centering\small\setlength{\tabcolsep}{5pt}
\caption{The losses of Table~\ref{tab:results} as mean $\pm$ standard deviation over timelines (GPT models; S1--S6, three timelines each). Rows are grouped by model, which is both the memory system's internal model and the reader. For maintenance, the model is the one that built the store, whose internal model is replaced at session 30 by the other GPT model; negative values mean that fewer stored facts were lost with the replacement. Lower is better.}
\label{tab:results-ci}
\resizebox{\textwidth}{!}{%
\begin{tabular}{llcccc}\toprule
Model & Memory system & Compression $\downarrow$ & Interference $\downarrow$ & Revision $\downarrow$ & Maintenance $\downarrow$ \\
\midrule
gpt-4o-mini & Naive RAG & 0.00 $\pm$ 0.00 & 0.37 $\pm$ 0.19 & 0.57 $\pm$ 0.36 & 0.00 $\pm$ 0.00 \\
 & HippoRAG 2 & 0.00 $\pm$ 0.00 & 0.30 $\pm$ 0.16 & 0.70 $\pm$ 0.25 & 0.00 $\pm$ 0.00 \\
 & Self-managed & 0.76 $\pm$ 0.20 & 0.17 $\pm$ 0.30 & 0.50 $\pm$ 0.35 & -0.21 $\pm$ 0.30 \\
 & Summarize & 0.88 $\pm$ 0.25 & 0.15 $\pm$ 0.17 & 0.56 $\pm$ 0.34 & -0.03 $\pm$ 0.41 \\
 & Consolidating & 0.88 $\pm$ 0.24 & 0.16 $\pm$ 0.38 & 0.19 $\pm$ 0.21 & 0.07 $\pm$ 0.28 \\
 & A-Mem & 0.00 $\pm$ 0.00 & 0.48 $\pm$ 0.17 & 0.72 $\pm$ 0.24 & 0.00 $\pm$ 0.00 \\
 & MemoryOS & 0.00 $\pm$ 0.00 & 0.35 $\pm$ 0.16 & 0.87 $\pm$ 0.17 & 0.00 $\pm$ 0.00 \\
 & LightMem & 0.22 $\pm$ 0.15 & 0.31 $\pm$ 0.20 & 0.65 $\pm$ 0.24 & 0.00 $\pm$ 0.00 \\
 & SimpleMem & 0.08 $\pm$ 0.09 & 0.29 $\pm$ 0.19 & 0.43 $\pm$ 0.30 & 0.00 $\pm$ 0.00 \\
 & Mem0 & 0.10 $\pm$ 0.10 & 0.29 $\pm$ 0.21 & 0.87 $\pm$ 0.20 & -0.01 $\pm$ 0.04 \\
\midrule
gpt-6-luna & Naive RAG & 0.00 $\pm$ 0.00 & 0.37 $\pm$ 0.15 & 0.54 $\pm$ 0.36 & 0.00 $\pm$ 0.00 \\
 & HippoRAG 2 & 0.00 $\pm$ 0.00 & 0.27 $\pm$ 0.15 & 0.59 $\pm$ 0.24 & 0.00 $\pm$ 0.00 \\
 & Self-managed & 0.09 $\pm$ 0.08 & 0.03 $\pm$ 0.07 & 0.06 $\pm$ 0.13 & 0.18 $\pm$ 0.25 \\
 & Summarize & 0.56 $\pm$ 0.30 & 0.21 $\pm$ 0.25 & 0.13 $\pm$ 0.23 & 0.24 $\pm$ 0.50 \\
 & Consolidating & 0.97 $\pm$ 0.07 & 0.17 $\pm$ 0.29 & 0.04 $\pm$ 0.11 & -0.01 $\pm$ 0.13 \\
 & A-Mem & 0.00 $\pm$ 0.00 & 0.36 $\pm$ 0.13 & 0.74 $\pm$ 0.31 & 0.00 $\pm$ 0.00 \\
 & MemoryOS & 0.00 $\pm$ 0.00 & 0.22 $\pm$ 0.20 & 0.85 $\pm$ 0.21 & 0.00 $\pm$ 0.00 \\
 & LightMem & 0.01 $\pm$ 0.04 & 0.23 $\pm$ 0.17 & 0.52 $\pm$ 0.23 & 0.00 $\pm$ 0.00 \\
 & SimpleMem & 0.00 $\pm$ 0.00 & 0.11 $\pm$ 0.09 & 0.09 $\pm$ 0.15 & 0.00 $\pm$ 0.00 \\
 & Mem0 & 0.08 $\pm$ 0.11 & 0.18 $\pm$ 0.17 & 0.35 $\pm$ 0.21 & 0.01 $\pm$ 0.04 \\
\bottomrule\end{tabular}}
\end{table}

\subsection{Supplementary Evidence}\label{app:supplementary-panels}

\paragraph{Who ages.}\label{app:who-ages} Table~\ref{tab:who-ages} gives fact accuracy by memory system and model, including the models and the full-context reference that Table~\ref{tab:results} does not show.

\begin{table}[ht]\centering\small
\caption{Who ages: fact accuracy at session 50 (drop from session 8 in parentheses) by memory system and model. A stronger model removes the aging of full context and of self-managed memory, not that of retrieval or of rewrite-based consolidation.}
\label{tab:who-ages}
\begin{tabular}{lccccc}\toprule
System & gpt-4o-mini & gpt-5-nano & gpt-5.6-luna & gpt-6-luna & claude-haiku-5.5 \\\midrule
Full context & 0.08 (-0.85) & 0.32 (-0.65) & 0.79 (-0.20) & 0.94 (-0.06) & 0.97 (-0.00) \\
Naive RAG & 0.58 (-0.37) & 0.44 (-0.52) & 0.62 (-0.36) & 0.62 (-0.37) & 0.69 (-0.29) \\
Self-managed & 0.19 (-0.29) & 0.18 (-0.27) & 0.88 (-0.05) & 0.87 (-0.03) & 0.92 (-0.05) \\
Consolidating & 0.07 (-0.45) & 0.04 (-0.35) & 0.06 (-0.72) & 0.02 (-0.47) & 0.08 (-0.83) \\
\bottomrule\end{tabular}
\end{table}

\paragraph{Derived state of revised attributes.} Table~\ref{tab:derived-state} asks three questions about the same revised attributes as the number of versions grows: the current version, the version before it, and how many versions there have been. With Naive RAG the current and previous versions fall once the versions outnumber what one search returns, and the count is already low with six versions. With full context the current version stays high, while the count falls to near zero: state that must be recomputed from the whole update history ages even when that history is entirely in view.

\begin{table}[ht]\centering\small
\caption{Derived state of revised attributes: accuracy at session 200 by the number of versions per revised attribute (200-session builds, reader gpt-6-luna; Naive RAG on all 116 timelines, full context on two timelines per scenario). \emph{Current}: the latest version; \emph{Previous}: the version before it; \emph{Count}: how many versions so far. Never-updated controls stay at 0.96--1.00 in every arm.}
\vspace{2mm}
\label{tab:derived-state}
\begin{tabular}{lcccccc}\toprule
 & \multicolumn{3}{c}{Naive RAG} & \multicolumn{3}{c}{Full context} \\
\cmidrule(lr){2-4}\cmidrule(lr){5-7}
Versions & Current & Previous & Count & Current & Previous & Count \\\midrule
6 & 0.97 & 0.93 & 0.69 & 0.94 & 0.92 & 0.92 \\
10 & 0.96 & 0.82 & 0.46 & 0.97 & 0.89 & 0.61 \\
20 & 0.53 & 0.21 & 0.00 & 0.94 & 0.78 & 0.17 \\
30 & 0.32 & 0.08 & 0.00 & 0.83 & 0.64 & 0.06 \\
\bottomrule\end{tabular}
\end{table}

\paragraph{Use in the reader's actual context.} The use error $U$ of Appendix~\ref{app:attribution-stages} is measured on clean evidence. Table~\ref{tab:use-context} compares it with $U_{\mathrm{ctx}}$, the use error when the same evidence is placed in what the reader actually saw at P1. With clean evidence the reader almost never fails at either checkpoint. In its actual context, the use error grows by session 50 for the systems that extract or rewrite memory, but not for Naive RAG, whose context consists of raw rounds.

\begin{table}[ht]\centering\small
\caption{Use error with clean evidence ($U=1-\mathrm{Acc}_{P3}$) and with the evidence placed in the reader's actual context ($U_{\mathrm{ctx}}=1-\mathrm{Acc}_{P3ctx}$), at sessions 8 and 50.}
\vspace{2mm}
\label{tab:use-context}
\begin{tabular}{lcccc}\toprule
 & \multicolumn{2}{c}{$U$} & \multicolumn{2}{c}{$U_{\mathrm{ctx}}$} \\
\cmidrule(lr){2-3}\cmidrule(lr){4-5}
Memory system & s8 & s50 & s8 & s50 \\\midrule
Naive RAG & 0.01 & 0.02 & 0.02 & 0.02 \\
LightMem & 0.01 & 0.02 & 0.02 & 0.12 \\
Mem0 & 0.01 & 0.02 & 0.04 & 0.20 \\
Self-managed & 0.00 & 0.03 & 0.02 & 0.08 \\
Summarize & 0.00 & 0.03 & 0.02 & 0.13 \\
Consolidating & 0.00 & 0.03 & 0.03 & 0.14 \\
\bottomrule\end{tabular}
\end{table}


\section{Implementation Details}\label{app:implementation}

This section documents the benchmark's components for reproducibility. Metric definitions are in Appendix~\ref{app:scoring-defs}; this section covers running an experiment, the generator's configuration, the memory systems and their prompts, and the cost.

\subsection{Running an Experiment}\label{app:overview}

A run has three steps. (1)~\emph{Build}: the generator writes, for every timeline of a configuration, the history, the answer key of every checkpoint, the temporal DAG, and the position of every planted round; it calls no model, and a configuration reproduces its build byte for byte. (2)~\emph{Evaluate}: a system under test writes the history session by session, is frozen at each checkpoint, and answers that checkpoint's probes; the same entry point runs the diagnostic conditions (Appendix~\ref{app:attribution-ext}) and the lifecycle event. (3)~\emph{Report}: aging curves and losses are computed from the per-probe records. Every model call is cached on its full request, so a run can be replayed exactly, and the store of every checkpoint is archived, so another reader can be evaluated on the same stores without rebuilding them.

\subsection{Generator and Pressure Configuration}\label{app:generators}

Each probe family is generated by one motif of the temporal DAG, and each motif exposes the knobs below. All knobs are fields of a configuration file and can be overridden at build time.

\begin{center}
\small
\renewcommand{\arraystretch}{1.2}
\begin{tabular}{p{4.2cm}p{8.9cm}}
\toprule
\textbf{Knob} & \textbf{Effect} \\
\midrule
\texttt{n\_sessions}, \texttt{checkpoints} & Length of the history; sessions at which the probes are asked. Adding or removing checkpoints after the first does not change the history. \\
\texttt{facts}, \texttt{rivals}, \texttt{topical} & Facts per timeline; other owners' values for the same attribute per fact; same-topic rounds without a value per fact. \\
\texttt{revisions}, \texttt{rev\_len}, \texttt{rev\_controls} & Revised attributes; versions per attribute, evenly spread over the history; never-updated controls. \\
\texttt{rev\_probes} & Extra questions per revised attribute: the previous version, the number of versions. \\
\texttt{hops}, \texttt{depth}, \texttt{split\_hops}, \texttt{hop\_place} & Chains per timeline; hops per chain; each later hop split into an introduction round and a value round; chains written early, or spread so that hop $h$ is written at the session of version $h$. \\
\texttt{aggregations}, \texttt{agg\_members}, \texttt{agg\_near}, \texttt{agg\_topical}, \texttt{agg\_place} & Count units; members per unit; near-miss items; rounds naming the anchor but no member; members written early, or accumulating over the history so that the count grows. \\
\texttt{scale\_ref} & Scales the competing content with \texttt{n\_sessions}, keeping its density per session constant. \\
\texttt{pool\_ref} & Makes the arms of a dose experiment bind the same content units, so that probes pair across arms. \\
\bottomrule
\end{tabular}
\renewcommand{\arraystretch}{1.0}
\end{center}

\noindent Three families of configurations are released (Table~\ref{tab:configs}). The 50- and 200-session configurations hold 116 timelines over S1--S6; dose arms that need scarce units (many members, deep chains) hold fewer.

\begin{table}[ht]
\centering
\small
\caption{Released configurations. All share 12 facts per timeline with 12 rivals and 6 same-topic rounds each, 3 revised attributes with 2 controls, 3 chains of depth 3 with split hops, and one aggregation unit with 6 members, 6 near-miss items and 24 anchor rounds, except where noted.}
\label{tab:configs}
\renewcommand{\arraystretch}{1.25}
\resizebox{\textwidth}{!}{
\begin{tabular}{lccp{6.4cm}}
\toprule
\textbf{Configuration} & \textbf{Sessions} & \textbf{Checkpoints} & \textbf{Differences} \\
\midrule
50-session & 50 & 8--50 (5 or 10) & 24 versions per revised attribute: the competing content of 200 sessions written within 50. \\
200-session & 200 & 8--200 (10) & 40 versions per revised attribute. \\
200-session dose arms & 200 & 8--200 (10) & As 200-session, plus three accumulating count units and the previous-version and count questions; each arm changes one knob: versions (6--30), members (6--40), or hop depth (2--8, spread). \\
\bottomrule
\end{tabular}}
\renewcommand{\arraystretch}{1.0}
\end{table}

\subsection{Memory Systems and Prompts}\label{app:memory}

Every memory system writes the history in time order through its own pipeline, with its LLM calls routed to the internal model of the system under test (Table~\ref{tab:memsys}). At a checkpoint the store is frozen; systems whose retrieval changes their state (MemoryOS) are probed on a copy.

\begin{table}[ht]
\centering
\small
\caption{Memory systems. Unit: what one write receives. The reader of a system with a search step uses the ReAct prompt of Appendix~\ref{app:session-anatomy}; the in-context memories are placed in the prompt.}
\label{tab:memsys}
\renewcommand{\arraystretch}{1.2}
\resizebox{\textwidth}{!}{
\begin{tabular}{llp{6.2cm}l}
\toprule
\textbf{System} & \textbf{Unit} & \textbf{What the internal model does} & \textbf{Embedding} \\
\midrule
Naive RAG & round & --- & text-embedding-3-small \\
HippoRAG~2 & round & entity and triple extraction; fact filtering at retrieval & NV-Embed-v2 \\
Self-managed & session & adds, rewrites and deletes notes in a notebook & --- \\
Summarize & session & rewrites the memory and the new session into a new memory & --- \\
Consolidating & session & keeps the latest session verbatim and folds older ones into a summary & --- \\
A-Mem & turn & builds a note per turn and evolves related notes & all-MiniLM-L6-v2 \\
MemoryOS & round & pages, topic summaries, user profile and knowledge & all-MiniLM-L6-v2 \\
LightMem & round & extracts memories from pre-compressed turns & all-MiniLM-L6-v2 \\
SimpleMem & window & extracts memory units; plans and reflects at retrieval & Qwen3-Embedding-0.6B \\
Mem0 & round & extracts facts and decides to add, update or delete & text-embedding-3-small \\
\bottomrule
\end{tabular}}
\renewcommand{\arraystretch}{1.0}
\end{table}

\noindent\textbf{Answer prompts.} Memories placed in the prompt and the full-context baseline answer in one call:

\begin{promptbox}{Answering from a memory placed in the prompt (Self-managed, Summarize, Consolidating)}
You are a personal assistant. Below is your memory of your past conversations with the user, written by you after each conversation.\\[4pt]
\textless{}the memory: the notebook's notes / the summary (+ raw window)\textgreater{}\\[4pt]
Answer the user's question from this memory. If information was updated, use the most recent version. The final answer must be ONLY the requested name or number (for a 'how many' question, only the number): no sentence, no explanation, no list of candidates.
\end{promptbox}

\begin{promptbox}{Answering from the full history (full context)}
You are a personal assistant. Below is the complete record of your past conversations with the user, each round with its date.\\[4pt]
\textless{}every round written before the checkpoint: {[}date{]} user: ...\\
assistant: ...\textgreater{}\\[4pt]
Answer the user's question from this record. If information was updated, use the most recent version. The final answer must be ONLY the requested name or number (for a 'how many' question, only the number): no sentence, no explanation, no list of candidates.
\end{promptbox}

\noindent\textbf{Self-managed memory.} The notebook is the only memory written by the tested model itself. After every session, one call receives the current notebook and that session's conversation and returns operations, applied in order (an operation on an unknown id is skipped). The prompt is identical for every model:

\begin{promptbox}{Self-managed: notebook update after every session (same for every model)}
You are a personal assistant who keeps a notebook about the user to help them in future conversations. The\\
notebook is all you will remember: the conversations themselves will be gone.\\[4pt]
NOTEBOOK (id: note):\\
\textless{}numbered notes, one per line: {[}id{]} note; (empty) at the first session\textgreater{}\\
Below is today's conversation. Update the notebook: add what you may need later about the user (their own things and\\
choices, names, numbers, plans, the people around them, and anything that changed), rewrite notes that changed, and\\
delete what is no longer true. Note what belongs to other people only as theirs. Keep each note short and\\
self-contained; add the date when it matters.\\[4pt]
CONVERSATION:\\
\textless{}Conversation on \textless{}date\textgreater{}:\\
user: ...\\
assistant: ...  (the whole session)\textgreater{}\\[4pt]
Return JSON \{"ops": {[}\{"op": "add", "text": "..."\}, \{"op": "update", "id": \textless{}id\textgreater{}, "text": "..."\}, \{"op": "delete", "id": \textless{}id\textgreater{}\}{]}\}
\end{promptbox}

\noindent\textbf{Summarize and Consolidating.} Both compact the history with the prompt below, led by the scenario of the build: its persona and content range are filled verbatim from the scenario definition, without reference to the probes. Summarize rewrites the memory and the new session into a new memory after every session; Consolidating keeps the latest session verbatim and folds older ones into the summary. The word budget $N$ is 2\% of the words of the history the summary stands for, within $[50, 4000]$; an output over $1.1N$ words is rewritten once with the same prompt, then cut. For S2, the persona is \emph{someone whose lifestyle assistant tracks their budgets, diet, preferred stores and weekly schedule}, and the content hints are personal budgets and spending limits by category; diet: allergies, restrictions and favourite foods; preferred stores, brands and where the user buys things; schedule: classes, appointments, recurring activities and their venues.

\begin{promptbox}{Summarize and Consolidating: compaction prompt}
You are a personal assistant's memory manager. The user is \textless{}persona of the scenario\textgreater{}. Below are your notes and conversations with the user.\\
Rewrite them as a concise memory. You MUST preserve ALL of the following verbatim:\\
\textless{}one line per content hint: - Everything about \textless{}hint\textgreater{}\textgreater{}\\
- Every other named person, place, organisation, product or thing the user mentions, and what it is to the user\\
- Every number, amount and date, with what it refers to\\
- Every item of a list the user keeps (each one separately)\\
- Every update: when something changed, the most recent value with its date\\
Write only what the text states; do not add any detail that is not in it. Be concise but complete.\\
Write at most \textless{}N\textgreater{} words.\\[4pt]
DOCUMENT:\\
\textless{}previous memory, then the new session(s)\textgreater{}\\[4pt]
SUMMARY:
\end{promptbox}

\noindent\textbf{Third-party systems.} Mem0 (1.0.11), LightMem (8449d57), SimpleMem (db80b6a), HippoRAG~2 (2.0.0a4), MemoryOS (587ed77) and A-Mem (b31a1dd) run with their own prompts and default settings. Changes needed to run them:
\begin{itemize}[leftmargin=*,nosep]
  \item Mem0: one conversation round per \texttt{add()}; the date its prompts give as today is fixed, so that cached calls stay valid.
  \item LightMem: the configuration of its LongMemEval script (LLMLingua-2 pre-compression, user-only extraction); eager attention in the segmenter; a failed pre-compression raises an error instead of silently keeping the raw text.
  \item HippoRAG~2: one passage per round (whole sessions overflow its entity-extraction prompt); an entity answer given as a bare JSON list is also parsed.
  \item SimpleMem: code defaults, with parallel retrieval off for a deterministic order.
  \item MemoryOS and A-Mem: code defaults; A-Mem writes one note per turn, as in its LoCoMo driver.
\end{itemize}

\subsection{Cost of Experiments}\label{app:cost}

Table~\ref{tab:cost} reports the API cost of one main-experiment cell (S1--S6, three timelines each, ten checkpoints). Calls answered from the shared cache are not charged, so the figures are lower bounds; local models (embeddings, LightMem's pre-compression) are not included.

\begin{table}[ht]
\centering
\small
\caption{API cost (USD) of one cell of the main experiments on 18 timelines.}
\vspace{2mm}
\label{tab:cost}
\begin{tabular}{lccccc}
\toprule
\textbf{Model} & \makecell{Naive RAG} & \makecell{HippoRAG-2} & \makecell{Self-managed} & \makecell{Summarize} & \makecell{Consolidating}  \\
\midrule
gpt-4o-mini & \$2.0 & \$4.4 & \$1.6 & \$1.4 & \$2.4 \\
gpt-6-luna & \$1.3 & \$5.2 & \$3.3 & \$3.9 & \$4.8  \\
\midrule
\textbf{Model} &  \makecell{A-Mem} & \makecell{MemoryOS} & \makecell{LightMem} & \makecell{SimpleMem} & \makecell{Mem0}  \\
\midrule
gpt-4o-mini & \$13.6 & \$13.7 & \$1.0 & \$14.5 & \$4.2 \\
gpt-6-luna &  \$18.0 & \$5.8 & \$1.1 & \$8.5 & \$6.0  \\
\bottomrule
\end{tabular}
\end{table}

\clearpage
\section{Case Studies}\label{app:casestudy}

This section walks through two probes of S2 in detail. \S\ref{app:compounding-error} follows one wrong answer given by six memory systems through the diagnostic conditions and into what each system kept; \S\ref{app:memory-example} follows one stored fact through ten checkpoints of two in-context memories. All excerpts are verbatim; numbers in brackets are session indices.

\subsection{Same Wrong Answer, Different Repairs}\label{app:compounding-error}

\begin{promptbox}{Multi-hop probe, depth 2 (S2, asked at session 50)}
\textbf{Q:} What banking service does my budget app integrate with? \quad \textbf{gold:} ZephyrBank\\[2pt]
\textbf{User's chain:} [6] My budget app is named PennyTrack. [7] Hey, I just found out about this cool service called PennyTrack that helps manage expenses. / I heard they integrate with a banking service known as ZephyrBank.\\[2pt]
\textbf{Distractor chain (another owner):} [34] Alex's finance tracker is named CoinWatch. [40] I'm looking into this service called CoinWatch. / It integrates with a banking service known as StarlingBank.
\end{promptbox}

\noindent At session 50, six of the seven memory systems in the attribution experiment (gpt-6-luna as internal model and reader) answer \emph{StarlingBank}, the value of the other owner's chain. The diagnostic conditions of Appendix~\ref{app:attribution-ext} locate the failure at three different stages (Table~\ref{tab:case-ladder}).

\begin{table}[ht]
\centering
\small
\caption{One wrong answer, three stages. Gold seen: the gold value is among the items the reader retrieved (systems with a search step) or in the memory text (in-context memories) at P1. Recovered at: the first condition under which the answer is correct.}
\label{tab:case-ladder}
\begin{tabular}{llccl}
\toprule
\textbf{System} & \textbf{P1 answer} & \textbf{Gold seen at P1} & \textbf{Recovered at} & \textbf{Stage} \\
\midrule
Naive RAG & StarlingBank & yes & P2 & retrieval \\
LightMem & StarlingBank & yes & P2 & retrieval \\
MemoryOS & StarlingBank & no & P2 & retrieval \\
Mem0 & StarlingBank & no & P3 & write \\
Summarize & StarlingBank & no & P3 & write \\
Consolidating & StarlingBank & no & P3 & write \\
Self-managed & ZephyrBank & yes & --- & --- \\
\bottomrule
\end{tabular}
\end{table}

\noindent What each system had in front of it explains the three stages.

\begin{promptbox}{Retrieval stage: Naive RAG at P1 (one search)}
\textbf{Search:} budget app banking service integrates with\\[2pt]
\textbf{Top results, in rank order (3 of 10 shown):}\\
{[}40{]} user: It integrates with a banking service known as StarlingBank.\\
{[}7{]} user: I heard they integrate with a banking service known as ZephyrBank.\\
{[}6{]} user: My budget app is named PennyTrack.\\[2pt]
\textbf{Not retrieved:} [34] Alex's finance tracker is named CoinWatch. [40] I'm looking into this service called CoinWatch.\\[2pt]
\textbf{Answer:} StarlingBank
\end{promptbox}

\noindent The store holds every round, and the search returns both values, but not the rounds that say whose service CoinWatch is; the StarlingBank round, whose subject is only ``it'', reads as if it continued the user's own chain. Handing over the rounds of both chains (P2) recovers the answer: the repair is in retrieval.

\begin{promptbox}{Write stage: what Mem0 stored (4 of 373 items)}
The user's budget app is named PennyTrack, a service that helps manage expenses.\\
Asked where ZephyrBank is based.\\
Alex's finance tracker is named CoinWatch.\\
It integrates with a banking service known as StarlingBank.
\end{promptbox}

\noindent Mem0 kept both apps, but no item ties ZephyrBank to PennyTrack, while the StarlingBank item survives without its subject. Even when these items are handed over (P2), the reader answers StarlingBank; only the raw rounds (P3) recover the answer: the repair is in what Mem0 writes.

\begin{promptbox}{Write stage: what the in-context memories kept at session 50}
\textbf{Summarize} (1{,}769 words): PennyTrack tracks expenses. [\ldots] Alex's finance tracker: CoinWatch; user looking into CoinWatch, which integrates with StarlingBank, headquartered in Lumenville.\\[2pt]
\textbf{Consolidating} (summary of 1{,}713 words): CoinWatch is Alex's finance tracker and integrates with StarlingBank. \emph{(No mention of PennyTrack or ZephyrBank.)}\\[2pt]
\textbf{Self-managed} (answers correctly): ZephyrBank, which integrates with PennyTrack, is headquartered in Novaville. [\ldots] Alex uses a finance tracker called CoinWatch. User is looking into CoinWatch; StarlingBank integrates with CoinWatch and is headquartered in Lumenville.
\end{promptbox}

\noindent Summarize kept the user's app but dropped ZephyrBank, and recorded the distractor's integration as the user's own interest; Consolidating kept only the distractor chain. In both, the gold is no longer in memory, so no reader can recover it: the repair is in what is kept. The notebook kept both chains with their owners. The symptom is the same for six systems; the repairs are a better retrieval for three, and a better write for the other three.

\subsection{A Stored Fact Lost by an In-Context Memory}\label{app:memory-example}

The fact below is stated once, in session 4, and never mentioned again by the user. Other people's cosmetics brands follow over the history, for example:

\begin{promptbox}{Fact probe (S2) and some of its rivals}
\textbf{Q:} Could you remind me which cosmetics brand I have a particular fondness for? \quad \textbf{gold:} VelvetEssence\\[2pt]
\textbf{Evidence:} [4] I'm putting together a shopping list for skincare products, and since I'm loyal to VelvetEssence, I need to make sure I have all their essentials. [\ldots]\\[2pt]
\textbf{Rivals (12), e.g.:} [9] I heard my neighbor Sarah got hooked on MoonAura cosmetics after trying their foundation. [\ldots] [26] Nathan, my college roommate, is a fan of LushLuxe because of their exotic scents. He's convinced me to try some when I can. [31] My colleague, Emma, uses SparkSilhouette for her cosmetics because she loves their radiant finish. [\ldots]
\end{promptbox}

\noindent Table~\ref{tab:case-fact} follows the fact through two in-context memories: Summarize with gpt-6-luna and Self-managed with gpt-4o-mini.

\begin{table}[ht]
\centering
\small
\caption{One fact at every checkpoint. Size: words of the memory text (Summarize) or number of notes (Self-managed). In memory: the gold value occurs in the memory. Answer: the reader's answer (same model as the internal model).}
\label{tab:case-fact}
\resizebox{\textwidth}{!}{
\begin{tabular}{rccl ccl}
\toprule
 & \multicolumn{3}{c}{\textbf{Summarize} (gpt-6-luna)} & \multicolumn{3}{c}{\textbf{Self-managed} (gpt-4o-mini)} \\
\cmidrule(lr){2-4}\cmidrule(lr){5-7}
\textbf{Session} & \textbf{Size} & \textbf{In memory} & \textbf{Answer} & \textbf{Size} & \textbf{In memory} & \textbf{Answer} \\
\midrule
8  & 340   & yes & VelvetEssence & 24  & yes & VelvetEssence \\
12 & 491   & yes & VelvetEssence & 44  & yes & MoonAura \\
16 & 717   & yes & VelvetEssence & 59  & yes & MoonAura \\
20 & 893   & yes & VelvetEssence & 71  & yes & MoonAura \\
25 & 1{,}057 & yes & VelvetEssence & 5   & no  & I'm sorry, but I don't have any information [\ldots] \\
30 & 1{,}255 & yes & VelvetEssence & 28  & no  & You haven't mentioned a specific cosmetics brand [\ldots] \\
35 & 1{,}470 & no  & Not specified & 57  & no  & SparkSilhouette \\
40 & 1{,}683 & no  & Not specified & 74  & no  & LushLuxe \\
45 & 1{,}776 & no  & PrismAura     & 100 & no  & LushLuxe \\
50 & 1{,}943 & no  & I don't have that information. & 126 & no & LushLuxe \\
\bottomrule
\end{tabular}}
\end{table}

\begin{promptbox}{Summarize (gpt-6-luna): the fact drifts to the end of a growing summary}
\textbf{Session 8} (340 words, the line at word 136): **Stores/brands:** [\ldots] Preferred clothing boutique: SilkenThread; exclusive line: VelvetFrost; flagship jacket: Midnight Whisper. Loyal to skincare brand VelvetEssence. Wants to check essentials and sales at preferred stores. [\ldots]\\[2pt]
\textbf{Session 30} (1{,}255 words, the line at word 1{,}164): [\ldots] Preferred boutique SilkenThread; exclusive line VelvetFrost; flagship jacket Midnight Whisper. Loyal skincare brand VelvetEssence; wants to check essentials/sales at preferred stores. [\ldots]\\[2pt]
\textbf{Sessions 31--35}: every rewrite exceeds the word budget even after the second pass and is cut from the end, by 159 to 630 words.\\[2pt]
\textbf{Session 50} (1{,}943 words): the only skincare line left is about another person: Niece Olivia drinks 2825 milliliters daily for skincare and believes it makes a noticeable difference.
\end{promptbox}

\noindent As the summary grows, the line about the brand moves from 40\% to 93\% of its length by session 30, toward the end that the budget cuts. It disappears in the rewrites between sessions 30 and 35, all of which were cut. A fact the summary still holds becomes fragile once it sits near the end of a text that grows faster than its budget.

\begin{promptbox}{Self-managed (gpt-4o-mini): the notebook rewrites the fact, then deletes itself}
\textbf{Session 8} (24 notes): [19] User is putting together a shopping list for skincare products from their preferred store, VelvetEssence.\\
\textbf{Sessions 12--20}: note 19 is still present, next to [44] User's neighbor Sarah got hooked on MoonAura cosmetics after trying their foundation. The reader answers MoonAura.\\[2pt]
\textbf{Update after session 24}: one update deletes 86 notes; 5 notes remain.\\[2pt]
\textbf{Session 50} (126 notes), e.g.: [109] User's college roommate is named Nathan, who is a fan of LushLuxe because of their exotic scents and has convinced the user to try some. [130] User's colleague Emma uses SparkSilhouette for her cosmetics because she loves their radiant finish.
\end{promptbox}

\noindent The notebook loses the fact in two steps. At write time, the brand becomes a ``store'', so that from session 12 the only note that mentions \emph{cosmetics} is a rival's and the reader answers with it although the gold is still in memory. After session 24, a single update deletes almost the whole notebook. The notes written afterwards come from later sessions, which hold only rivals; one of them is phrased as something the user was persuaded to try, and the reader answers with it.

\noindent Both memories held the fact when its writing phase ended, and both lost it while rewriting their own contents. Summarize loses it silently and the reader declines to answer; the notebook loses it to a rival and the reader answers wrongly with confidence.
 that rival.

\section{Evaluation Card}\label{app:eval-card}

\textbf{What \textsc{AgingBench} reports.}
For each system under test (a memory system with its internal model), \textsc{AgingBench} reports aging curves of a fixed probe set at every checkpoint of a timeline, one loss per aging mechanism (compression, interference, revision, maintenance), and a component-aware diagnosis (\S\ref{sec:attribution}) that points to the write, retrieval, or utilization stage of the memory pipeline. Histories are generated deterministically from frozen content libraries (Appendix~\ref{app:curation}), so every build is reproducible from its configuration, and every probe is scored by rule against its answer key.

\textbf{Scope of attribution claims.}
We treat the component-aware diagnosis as \emph{counterfactual diagnosis under paired controls}, not as exact additive causal decomposition. The conditions (P1, P2, P3, P3gold, P3ctx) yield response profiles that we interpret as diagnostic profiles pointing to a candidate stage; we do not claim unique causal identification. The oracles of P2 and P3 are relevance oracles: they hand over the answer together with its competitors, without indicating which is which. Where the additive accounting requires the ladder to be monotone and a cell violates it, we report the cell as a \emph{diagnostic anomaly} rather than as a stage magnitude. For memories without a retrieval step (memories placed whole in the prompt), P2 is not run and write and retrieval errors are reported jointly.

\textbf{Intended use.}
Two regimes: (i)~\emph{evaluating a memory system or a model}: a new system under test is registered once and run on the released builds, which yields its aging curves and mechanism losses next to the systems evaluated here; and (ii)~\emph{diagnostic analysis} (\S\ref{sec:attribution}) that identifies which mechanism binds a given system and which stage a repair should target.


\clearpage

\section{Broader Discussion}\label{sec:discussion}

\textbf{Implications for long-lived agent system design.}
Three design observations follow from our results. First, memory design and model capability interact: a stronger model removes part of the aging of some memories, such as full context or a self-managed notebook, but not that of retrieval or of rewrite-based consolidation, so investment in a stronger model should be paired with an assessment of which component binds. Second, monitoring answers alone can miss aging: a memory that loses a fact may lead the agent to decline, or to answer confidently with a competitor, and only probes of what the agent should still know expose the difference. Third, memory \emph{curation} is an active operation with its own failure modes: rewriting, consolidating and deleting stored content can remove knowledge the agent already had, so the operations that keep a memory within its budget deserve the same scrutiny as the operations that write it.

\textbf{Connection to task horizons.}
If the \emph{task horizon} an agent can reliably handle is a north-star metric for agent progress, \textsc{AgingBench} supplies the longitudinal component that one-shot evaluations cannot. Its aging curves describe how reliability erodes as the history grows and as competing content accumulates, and its pressure knobs vary these axes one at a time. In the configurations we tested, a stronger model closes part of this erosion but not all of it. Extending the usable task horizon therefore requires intervening on the specific mechanism that binds the regime in question, whether compression of what is written, interference as similar content accumulates, revision as user state evolves, or maintenance around lifecycle events; single-axis improvements are unlikely to yield deployment-stable agents.

\textbf{Aging as a runtime control problem.}
A long-lived agent makes continuous decisions about what to write, what to compress, what to retrieve, and when to reorganize its memory, mirroring the deliberate practices humans use to maintain cognitive function with age. Read this way, each memory system in our evaluation is a point in a control-policy space, and its per-mechanism aging curves are the closed-loop evaluation of that policy. The per-checkpoint signals \textsc{AgingBench} emits could in principle drive such decisions at run time, which makes it a natural testbed for runtime-control studies within agent lifespan engineering.
State that must be recomputed from a whole update history, such as how many times a value changed or what it was before, ages even when the full history is in view. This suggests that text memory may be the wrong abstraction for such state, and that keeping it in an explicit, structured form updated at write time is a natural direction for memory design.

\textbf{LLM Declaration.} LLMs are central to this work as the research objective under evaluation. They appear methodologically as the experimental subject (multiple LLMs across families) and as components of the evaluated pipeline (the memory management; the tool-calling loop in ReAct loop for agent). Scoring does not use LLM-as-judge: metrics are computed by deterministic procedures over gold ground truth carried through the temporal dependency DAG. The authors additionally used LLMs for language refinement editing (e.g., grammar, spelling, word choice), code implementation, and experiment conducting; this did not influence methodology design, claims, or originality.

\textbf{Limitations and the open frontier.}
\textsc{AgingBench} contributes a mechanism-level vocabulary for agent lifespan engineering, covering four mechanisms (compression, interference, revision, maintenance), along with a reproducible protocol that measures them over controlled history lengths and diagnoses aging at specific pipeline stages rather than against aggregate model quality. Its scope has limits. The histories are synthetic: facts are planted into scenario-themed conversations, and base conversations recur across the timelines of a build, so timelines are not fully independent. Most probes ask for a name or a number; the task form covers part of the probes. The content libraries were authored and checked with particular model families, which may favor some readers. A small share of items and scores remains flawed despite the checks. Lifecycle events can be extended in future work to more types of operations on stored memory, such as flushes and migrations. The open frontier is anchoring this vocabulary to production deployment telemetry to verify at real-user timescales that the mechanisms characterized here compound over weeks-long deployments; we hope the community can extend this diagnostic lens to deployed systems.


\newpage
\section*{NeurIPS Paper Checklist}

\begin{enumerate}

\item {\bf Claims}
    \item[] Question: Do the main claims made in the abstract and introduction accurately reflect the paper's contributions and scope?
    \item[] Answer: \answerYes{}
    \item[] Justification: We have summarized the position and key contributions of the paper in the abstract and introduction parts.

\item {\bf Limitations}
    \item[] Question: Does the paper discuss the limitations of the work performed by the authors?
    \item[] Answer: \answerYes{}
    \item[] Justification: The limitations are discussed in Appendix~\ref{sec:discussion}.

\item {\bf Theory assumptions and proofs}
    \item[] Question: For each theoretical result, does the paper provide the full set of assumptions and a complete (and correct) proof?
    \item[] Answer: \answerNA{}
    \item[] Justification: There are no theoretical results in this paper.

    \item {\bf Experimental result reproducibility}
    \item[] Question: Does the paper fully disclose all the information needed to reproduce the main experimental results of the paper to the extent that it affects the main claims and/or conclusions of the paper (regardless of whether the code and data are provided or not)?
    \item[] Answer: \answerYes{}
    \item[] Justification: We introduced the experiment and implementation details in Appendices~\ref{app:setup}, \ref{app:task-examples}, \ref{app:implementation}, and~\ref{app:multiseed}.

\item {\bf Open access to data and code}
    \item[] Question: Does the paper provide open access to the data and code, with sufficient instructions to faithfully reproduce the main experimental results, as described in supplemental material?
    \item[] Answer: \answerYes{}
    \item[] Justification: The code can be accessed by our project page at \url{https://agingbench.github.io}, with details introduced in Appendix~\ref{app:implementation}.

\item {\bf Experimental setting/details}
    \item[] Question: Does the paper specify all the training and test details (e.g., data splits, hyperparameters, how they were chosen, type of optimizer) necessary to understand the results?
    \item[] Answer: \answerYes{}
    \item[] Justification: The experimental setup and configurations are documented in Appendices~\ref{app:setup} and~\ref{app:memory}.

\item {\bf Experiment statistical significance}
    \item[] Question: Does the paper report error bars suitably and correctly defined or other appropriate information about the statistical significance of the experiments?
    \item[] Answer: \answerYes{}
    \item[] Justification: The multi-seed validation results are reported in Appendix~\ref{app:multiseed}.

\item {\bf Experiments compute resources}
    \item[] Question: For each experiment, does the paper provide sufficient information on the computer resources (type of compute workers, memory, time of execution) needed to reproduce the experiments?
    \item[] Answer: \answerYes{}
    \item[] Justification: The compute resources are summarized in Appendix~\ref{app:cost}.

\item {\bf Code of ethics}
    \item[] Question: Does the research conducted in the paper conform, in every respect, with the NeurIPS Code of Ethics \url{https://neurips.cc/public/EthicsGuidelines}?
    \item[] Answer: \answerYes{}
    \item[] Justification: The work conforms with the NeurIPS Code of Ethics.

\item {\bf Broader impacts}
    \item[] Question: Does the paper discuss both potential positive societal impacts and negative societal impacts of the work performed?
    \item[] Answer: \answerYes{}
    \item[] Justification: The broader impacts are discussed in Appendix~\ref{sec:discussion}.

\item {\bf Safeguards}
    \item[] Question: Does the paper describe safeguards that have been put in place for responsible release of data or models that have a high risk for misuse (e.g., pre-trained language models, image generators, or scraped datasets)?
    \item[] Answer: \answerNA{}
    \item[] Justification: This paper does not release new models.

\item {\bf Licenses for existing assets}
    \item[] Question: Are the creators or original owners of assets (e.g., code, data, models), used in the paper, properly credited and are the license and terms of use explicitly mentioned and properly respected?
    \item[] Answer: \answerYes{}
    \item[] Justification: All assets used in the paper are cited in Appendix~\ref{app:setup} and Appendix~\ref{app:related}.

\item {\bf New assets}
    \item[] Question: Are new assets introduced in the paper well documented and is the documentation provided alongside the assets?
    \item[] Answer: \answerYes{}
    \item[] Justification: The released benchmark is documented in Appendices~\ref{app:task-examples} and~\ref{app:implementation}, with intended use described in the Evaluation Card (Appendix~\ref{app:eval-card}).

\item {\bf Crowdsourcing and research with human subjects}
    \item[] Question: For crowdsourcing experiments and research with human subjects, does the paper include the full text of instructions given to participants and screenshots, if applicable, as well as details about compensation (if any)?
    \item[] Answer: \answerNA{}
    \item[] Justification: The paper is not about crowdsourcing experiments or research with human subjects.

\item {\bf Institutional review board (IRB) approvals or equivalent for research with human subjects}
    \item[] Question: Does the paper describe potential risks incurred by study participants, whether such risks were disclosed to the subjects, and whether Institutional Review Board (IRB) approvals (or an equivalent approval/review based on the requirements of your country or institution) were obtained?
    \item[] Answer: \answerNA{}
    \item[] Justification: The paper is not about research with human subjects.

\item {\bf Declaration of LLM usage}
    \item[] Question: Does the paper describe the usage of LLMs if it is an important, original, or non-standard component of the core methods in this research? Note that if the LLM is used only for writing, editing, or formatting purposes and does \emph{not} impact the core methodology, scientific rigor, or originality of the research, declaration is not required.
    \item[] Answer: \answerYes{}
    \item[] Justification: The LLM usage is described in the LLM Declaration paragraph of Appendix~\ref{sec:discussion}.
    \item[] Guidelines:
    \begin{itemize}
        \item The answer \answerNA{} means that the core method development in this research does not invade LLMs as any important, original, or non-standard components.
        \item Please refer to our LLM policy in the NeurIPS handbook for what should or should not be described.
    \end{itemize}

\end{enumerate}

\end{document}